\documentclass{article}

\usepackage[preprint]{neurips_2026}

\usepackage{algorithm}
\usepackage{algpseudocode}
\usepackage{amsfonts}       % blackboard math symbols
\usepackage{amsmath}
\usepackage{amssymb}
\usepackage{amsthm}
\usepackage{bbm}
\usepackage{booktabs}       % professional-quality tables
\usepackage[table]{xcolor}
\usepackage[T1]{fontenc}    % use 8-bit T1 fonts
\usepackage{graphicx}
\usepackage{hyperref}       % hyperlinks
\usepackage[utf8]{inputenc} % allow utf-8 input
\usepackage{makecell}
\usepackage{mathtools}
\usepackage{mdframed}
\usepackage{microtype}      % microtypography
\usepackage{multicol}
\usepackage{multirow}
\usepackage{nicefrac}       % compact symbols for 1/2, etc.
\usepackage{subcaption}
\usepackage[most]{tcolorbox}
\usepackage{tikz,times}
\usepackage{url}            % simple URL typesetting
\usepackage{wrapfig}        % wrap text around figures
\usepackage{colortbl}

\usetikzlibrary{arrows,intersections}
\usetikzlibrary{mindmap,backgrounds}
\usetikzlibrary{positioning}
\usetikzlibrary{snakes}

\tikzstyle{block} = [draw, rectangle, minimum height=3em, minimum width=5em, align=left]
\tikzstyle{pinstyle} = [pin edge={to-,thin,black}]

\newtheorem{theorem}{Theorem}
\newtheorem{definition}{Definition}
\newtheorem{proposition}{Proposition}
\newtheorem{lemma}{Lemma}

\newtheorem{remark}{Remark}

\usepackage{pifont}         % \ding symbols
\usepackage{float}
\usepackage{placeins}

\newcommand{\revision}[1]{#1}

\definecolor{pinktext}{RGB}{255,105,180}

\newtcolorbox{promptbox}[2][]{
  boxrule=0pt,
  colback=gray!5,
  colframe=gray!70,
  fontupper=\small,
  title={#2},
  coltitle=white,
  colbacktitle=darkgray,
  width=\linewidth,
  breakable,
}

\title{Calibration-Constrained Diagnostics for Evaluating LLM Self-Evaluation\thanks{This paper is an extended and substantially revised version of our workshop paper ``Can LLMs Reliably Evaluate Themselves? A Probabilistic VC Framework'', which was published as a non-archival paper at the NeurIPS 2025 Workshop on Bridging Language, Agent, and World Models for Reasoning and Planning (LAW 2025; \url{https://openreview.net/forum?id=CdxeuFBAUD}). The relation between the two versions is described at the end of the Introduction.}}

\author{%
  Jae Oh Woo \quad Mengdie Flora Wang \quad Rahul Ghosh \quad Baishali Chaudhury \quad Mun Young Kim \\
  AWS Generative AI Innovation Center \\
  \texttt{\{jaeohwoo, florawan, rahulgh, baishch, munkim\}@amazon.com}
}

\begin{document}

\maketitle

\begin{abstract}
We introduce a reusable evaluation artifact\footnote{\raggedright Code and data: \url{https://github.com/calibration-constrained-diagnostics/selfeval-diagnostics/}} for diagnosing whether LLM self-reported confidence can support confidence-gated act-or-defer decisions. The artifact is designed for evaluation rather than safety certification: it provides a black-box protocol, prompts, judge-reference labels, model outputs, reproduction scripts, and calibration-constrained diagnostic metrics for auditing LLM self-evaluation across math, factual, and commonsense reasoning tasks. Our central finding is that aggregate uncertainty summaries can produce deployment-misleading rankings: a model that appears strongest under standard aggregate calibration metrics can fail under category-level calibration constraints, while a model with worse aggregate calibration can exhibit stronger calibration-constrained behavior in the positive-margin region where high self-selection accuracy and tight calibration coexist. This reversal occurs because aggregate metrics can hide category-level failures: high-confidence errors and near-chance predictions cancel in averages. To expose this failure mode, we define PVC and C-PVC as category-level diagnostics over an accuracy-by-calibration surface: PVC counts categories whose self-evaluation accuracy exceeds a threshold, while C-PVC additionally requires category-level calibration error below a tolerance. We aggregate these diagnostics using Volume Under Surface (VUS) and a positive-margin restriction; these diagnostics are VC/fat-shattering motivated but are empirical evaluation proxies, not literal VC-dimension estimates. Across $13$ models and four benchmarks, the artifact shows weak confidence--correctness separation, stable diagnostic rankings under alternative categorizations and judge-label perturbations, and persistent positive-margin failures not captured by ECE or Brier score. Together, these results show that confidence-gated self-evaluation requires calibration-constrained, category-level diagnostics, and we provide a reusable artifact for applying them under black-box access.
\end{abstract}

\section{Introduction} \label{sec:introduction}

As large language models (LLMs) are increasingly embedded in autonomous reasoning pipelines, a practical question arises: when a model reports confidence, is that confidence informative for confidence-gated downstream decisions? Existing uncertainty quantification (UQ) tools (ECE~\citep{guo2017calibration}, Brier score, ensemble and semantic-entropy scorers~\citep{lmpolygraph, uqlm, benchmark_uq}, and confidence elicitation~\citep{xiong2024canllms, kadavath2022language}) report aggregate, distribution-level summaries. Aggregate summaries admit cancellation: a model placing half its mass near chance confidence and half at near-certainty can display low ECE yet be unreliable in both regimes, and can be \emph{preferred} over a model with better category-level calibrated behavior.

Rather than propose another scalar uncertainty score, we evaluate whether self-reported confidence remains meaningful after simultaneously imposing discrimination, per-category calibration, and positive-margin constraints. We define two category-level, VC/fat-shattering motivated diagnostic proxies on an accuracy-by-calibration surface: \emph{PVC} counts categories with self-evaluation accuracy $\mathrm{SEA}\ge\gamma$, and \emph{C-PVC} additionally requires per-category calibration error $\le\tau$. We aggregate across $(\gamma,\tau)$ via Volume Under Surface (VUS) and a \emph{positive-margin (PM)} restriction PM-VUS over $\gamma>\tau+1/2$, the region in which a standard uniform-convergence interpretation is available under explicit capacity-envelope assumptions. These are empirical evaluation proxies, not literal VC-dimension estimates.

\textbf{Contributions.}
\textbf{(i)} A reusable black-box \emph{evaluation protocol}: three-stage pairwise self-evaluation requiring only generation, binary self-selection, and a confidence report, applicable to closed/API-only models and agentic systems.
\textbf{(ii)} Calibration-constrained diagnostic \emph{metrics} (PVC, C-PVC, VUS, PM-VUS) that separate discrimination, per-category calibration, and positive-margin behavior, complementary to ECE/Brier and ensemble UQ scorers.
\textbf{(iii)} A \emph{controlled instantiation} on Math-360 (a $360$-problem diagnostic testbed in a supporting role, not a leaderboard) with cross-domain validation on TruthfulQA~\citep{lin-etal-2022-truthfulqa} and CommonsenseQA~\citep{talmor-etal-2019-commonsenseqa}, plus a MATH-500~\citep{huggingface2024math500} robustness check in the appendix (Appendix~\ref{appendix:artifact_card}).
\textbf{(iv)} \emph{Empirical finding}: across $13$ models (7B--32B plus Claude 3.5 Haiku) and four benchmarks, JiuZhang3.0-7B is best by ECE ($0.199$) and Brier ($0.049$) yet has PM-C-PVC-VUS only $0.019$; Qwen2.5-32B-Instruct has worse ECE ($0.260$) but $\approx 11\times$ higher PM-C-PVC-VUS ($0.209$): a deployment-relevant ranking reversal hidden by aggregate metrics. Across $17{,}420$ predictions, the mean correct--incorrect confidence gap is $0.005$; $21/52$ model-dataset pairs show a \emph{negative} gap.
\textbf{(v)} \emph{Robustness}: rankings are stable under $7$ alternative categorizations (Spearman $\ge 0.75$), $10$--$30\%$ synthetic judge-label perturbation, and unanimous-only re-analysis.

\textbf{Scope.} We do not claim to estimate VC dimension, that LLMs universally cannot self-evaluate, or that existing UQ tools should be replaced. We claim that aggregate uncertainty summaries can hide a deployment-relevant failure mode under black-box access and per-category calibration constraints.

\textbf{Relation to the workshop version.} This paper builds on our earlier workshop paper \citep{woo2025can}, published as a non-archival paper at the NeurIPS 2025 LAW workshop and posted as version 1 of this arXiv record, which introduced the PVC and C-PVC measures, the pairwise self-evaluation protocol, and the Math-360 testbed, and reported results for eleven 7--8B models. The present version extends that work in four ways: (i) it reframes PVC and C-PVC as VC/fat-shattering motivated empirical evaluation proxies rather than capacity estimates, and adds the positive-margin restriction PM-VUS with the corresponding uniform-convergence interpretation; (ii) it enlarges the evaluation to $13$ models, including Qwen2.5-32B-Instruct and Claude 3.5 Haiku, and centers the analysis on the ranking reversal between aggregate calibration metrics and calibration-constrained diagnostics; (iii) it adds robustness analyses under alternative categorizations, judge-label perturbation, and unanimous-only reference labels; and (iv) it releases the full evaluation artifact (protocol, prompts, reference labels, model outputs, and reproduction scripts).

\section{Background}\label{sec:background}

\paragraph{Classical VC dimension.}
For a binary hypothesis class $\mathcal{H}\subseteq\{h:\mathcal{X}\to\{0,1\}\}$, a set $\{x_1,\dots,x_n\}$ is \emph{shattered} by $\mathcal{H}$ if every labeling $(y_1,\dots,y_n)\in\{0,1\}^n$ is realized by some $h\in\mathcal{H}$; $\mathrm{VC}(\mathcal{H})$ is the size of the largest such set~\citep{vapnik1971theory, blumer1989learnability, shalev2014understanding}. If $\mathrm{VC}(\mathcal{H})=d$, standard uniform convergence gives, for any $\delta\in(0,1)$ and sample of size $n$,
\[
\sup_{h\in\mathcal{H}}\bigl|R(h)-\hat R(h)\bigr|\;\le\;C\sqrt{\frac{d+\log(1/\delta)}{n}}
\]
with probability $\ge 1-\delta$, where $R$ and $\hat R$ are the true and empirical risks. Classical VC theory is restricted to binary outputs and is inadequate for LLMs with probabilistic predictions.

\paragraph{Fat-shattering dimension.}
For bounded real-valued classes $\mathcal{F}\subseteq[0,1]^\mathcal{X}$ and scale $\alpha>0$, a set is $\alpha$-shattered if there exist thresholds $s_i$ such that for every sign pattern $\sigma\in\{\pm 1\}^n$ some $f\in\mathcal{F}$ satisfies $\sigma_i(f(x_i)-s_i)\ge\alpha$ for all $i$; $\mathrm{fat}_\alpha(\mathcal{F})$ is the largest such $n$~\citep{bartlett1995more, alon1997scale, bartlett1994fat}. For binary classes $\mathrm{fat}_\alpha(\mathcal{F})=\mathrm{VC}(\mathcal{F})$ on $\alpha\in(0,1)$ and zero for $\alpha\ge 1$. Fat-shattering is the scale-sensitive extension of VC that we use to motivate the positive-margin region.

\section{Methods}\label{sec:methods}

\paragraph{Convention: idealized vs.\ empirical.}
Throughout, $\mathrm{PVC}_\gamma(\mathcal{F})$ and $\mathrm{C\text{-}PVC}_\gamma^\tau(\mathcal{F})$ denote \emph{idealized} population-level capacity notions defined via shattering. Their \emph{empirical} counterparts (no hat for readability) are category-level black-box diagnostics computed from SEA and calibration error on observed categories. The acronym ``PVC'' is retained to reflect the motivating idealized capacity notion; all empirical PVC/C-PVC values reported in this paper should be read as category-count diagnostics over a fixed evaluation design, not as literal VC-dimension estimates.

\subsection{Idealized PVC and C-PVC}
\label{sec:pvc-cpvc}

We extend the classical VC framework to probabilistic predictors $f:\mathcal{X}\to\Delta(\{0,1\})$, where $\Delta(\{0,1\})$ is the set of distributions over the binary output space. Rather than requiring exact realizability, we require each labeling to be realized \emph{in probability}.

\begin{definition}[Idealized Probabilistic VC Capacity]\label{def:pvc}
Fix $\gamma\in(0,1]$. $\mathrm{PVC}_\gamma(\mathcal{F})$ is the largest $d$ such that some $\{x_1,\dots,x_d\}\subseteq\mathcal{X}$ admits, for every labeling $(y_1^*,\dots,y_d^*)\in\{0,1\}^d$, a predictor $f\in\mathcal{F}$ with $\mathbb{P}(f(x_i)=y_i^*)\ge\gamma$ for all $i$.
\end{definition}

PVC quantifies discriminative capacity but does not require confidence to match correctness. For high-stakes deployment, calibration is essential:

\begin{definition}[Idealized Calibration-aware PVC Capacity]\label{def:c-pvc}
Fix $\gamma\in(0,1]$ and $\tau\in[0,1)$. $\mathrm{C\text{-}PVC}_\gamma^\tau(\mathcal{F})$ is the largest $d$ such that some $\{x_1,\dots,x_d\}$ admits, for every labeling, a pair $(f,\hat{p})\in\mathcal{F}$ with
\[
\mathbb{P}(f(x_i)=y_i^*)\ge\gamma
\quad\text{and}\quad
\bigl|\hat{p}(x_i)-\mathbb{E}[\mathbf{1}\{f(x_i)=y_i^*\}]\bigr|\le\tau \;\;\text{for all}\; i.
\]
\end{definition}

Unlike population-level ECE~\citep{guo2017calibration}, C-PVC imposes calibration on \emph{fixed} inputs, preventing cancellation between over- and under-confidence errors.

\begin{figure}[!t]
    \centering
    \includegraphics[width=0.85\linewidth]{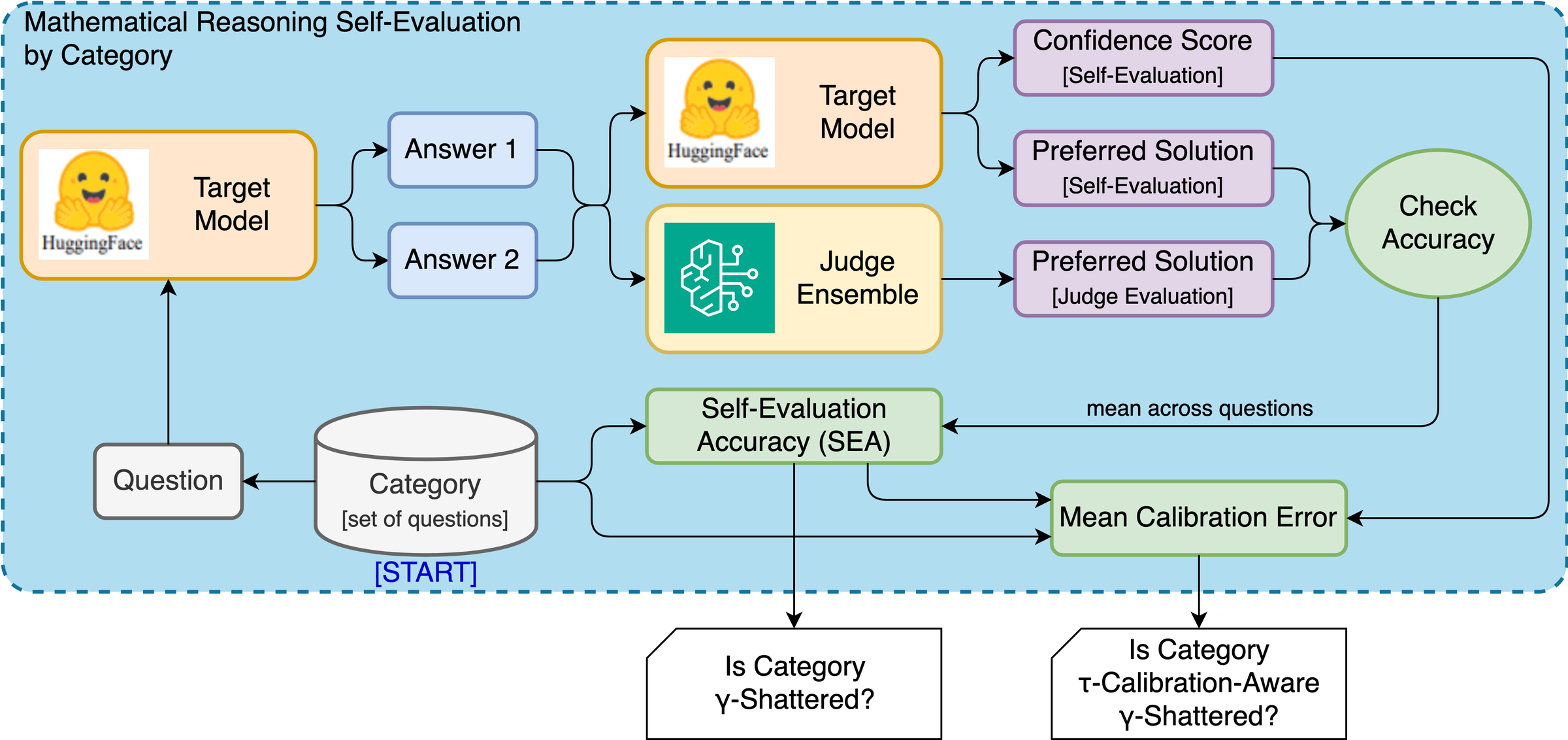}
    \caption{Self-evaluation diagnostic pipeline. The target model generates two candidate solutions, selects the preferred one with a confidence report, and is evaluated against a three-judge LLM reference label.}
    \label{fig:exp-protocol}
\end{figure}

\paragraph{LLMs as function classes.}
A pretrained LLM is not a single function but a \emph{family} of predictors induced by parameters, prompts, and stochastic decoding. Three perspectives motivate this view: (i) a PAC-Bayesian / parameter-level randomization view~\citep{mcallester2003simplified, germain2016pac}; (ii) a prompt-based input-conditioning view in which a prompt $p\in\Pi$ indexes a hypothesis $x\mapsto f_W([p;x])$~\citep{li2021prefix, lester2021power}; and (iii) a stochastic-decoding view that maps the model to a Gibbs-type probabilistic predictor~\citep{gal2015dropout}. Combining these, we fix pretrained parameters $W$ and vary prompts $\pi\in\Pi$ and decoding $\omega\in\Omega$: $\mathcal{F}(W)=\{f(\cdot;W,\pi,\omega):\pi\in\Pi,\omega\in\Omega\}$. This is the operational hypothesis class to which our idealized capacity notions are applied (Appendix~\ref{appendix:llm_function_class}).

\subsection{Measuring PVC and C-PVC through Self-Evaluation}\label{subsec:measuring_pvc_cpvc}

Direct shattering verification is intractable ($2^d$ labelings) and statistically unstable at the question level. We therefore operate at the \emph{category level}: semantic categories (e.g., Algebra, Geometry) act as distributional evaluation units, with performance estimated by aggregating over questions sampled i.i.d.\ from a category-specific distribution $\mathcal{D}_C$. Categories are coherent reasoning domains, and satisfying the diagnostic constraints at the category level provides evidence of systematic discriminative behavior across problem types.

\textbf{Evaluation protocol.}
Three stages (Figure~\ref{fig:exp-protocol}): (1) the target model generates two distinct solutions for each problem; (2) the target model selects the preferred solution and reports a confidence on $[0,100]$ (randomized presentation order, no reference-label access); (3) a three-judge LLM ensemble (Claude 3.7 Sonnet, Amazon Nova Premier, DeepSeek-R1) assigns a reference preference by majority vote. Pairwise comparison is the foundational structure of preference optimization~\citep{rafailov2023direct, ouyang2022training} and the de facto standard for LLM-based evaluation~\citep{zheng2023judging}. The judge ensemble is used as a \emph{stable comparative reference}, not as oracle ground truth; the diagnostic signal is the \emph{difference} between models under a fixed reference.

\begin{definition}[Reference-Agreement Self-Evaluation Accuracy and Calibration Error]\label{def:sea}
Let $\mathsf{Select}_f(q)$ denote the model's selected answer, $\hat{p}(q)$ its reported confidence, and $\mathsf{Judge}(q)$ the reference label. For category $C$ with question distribution $\mathcal{D}_C$,
\[
\mathrm{SEA}(f,C) \;=\; \mathbb{E}_{q\sim\mathcal{D}_C}\bigl[\mathbf{1}\{\mathsf{Select}_f(q)=\mathsf{Judge}(q)\}\bigr],
\]
\[
\mathrm{CalibError}(f,C) \;=\; \bigl|\mathbb{E}_{q\sim\mathcal{D}_C}[\hat{p}(q)]-\mathrm{SEA}(f,C)\bigr|.
\]
SEA measures \emph{agreement with the fixed judge-reference preference}, not absolute correctness; the same reference is applied across all compared models, so SEA differences reflect self-evaluation behavior rather than differences in the reference.
\end{definition}

A category is counted toward the empirical PVC proxy if $\mathrm{SEA}(f,C)\ge\gamma$, and toward the empirical C-PVC proxy if additionally $\mathrm{CalibError}(f,C)\le\tau$. These counts are \emph{operational diagnostics under the chosen prompt/decoding protocol}, not literal VC-dimension estimates and not certified lower bounds on the true hypothesis-class dimension. Full positioning and the theory-to-practice mapping (inputs $\mapsto$ categories, shattering $\mapsto$ SEA constraint, calibration $\mapsto$ CalibError constraint) are in Appendix~\ref{appendix:operational_proxy}.

\subsection{Aggregation: VUS and Positive-Margin (PM) Restriction}

To avoid threshold cherry-picking, we aggregate across $(\gamma,\tau)$ via
\begin{equation}
M\text{-}\mathrm{VUS}(\mathcal{F}) = \iint_{\mathcal{G}} M(\gamma, \tau, \mathcal{F})\,d\gamma\,d\tau,
\qquad \mathcal{G}=(0,1]\times[0,1),\; M\in\{\mathrm{PVC},\mathrm{C\text{-}PVC}\}.
\end{equation}
Theorem~\ref{thm:cpvc-generalization} applies only when $\gamma>\tau+1/2$, yielding a positive fat-shattering margin $\alpha=\gamma-\tau-1/2>0$. We thus report the \emph{positive-margin (PM)} variant
\begin{equation}
M\text{-}\mathrm{PM\text{-}VUS}(\mathcal{F}) = 2\iint_{\mathcal{G}^*}\tfrac{1}{\gamma-1/2}\, M(\gamma,\tau,\mathcal{F})\,d\gamma\,d\tau,
\qquad \mathcal{G}^*=\{(\gamma,\tau):\gamma>\tau+\tfrac{1}{2}\}.
\end{equation}
The $1/(\gamma-1/2)$ factor normalizes the $\tau$-range at each $\gamma$; the leading $2$ makes PM-VUS comparable to full-region VUS. Rankings agree with an \emph{unweighted} PM aggregation (Appendix~\ref{appendix:pg-vus-derivation}); ``PM'' denotes the positive-margin region only, not an empirical generalization guarantee.

\section{Theoretical Results}\label{sec:theoretical_results}

Two results motivate the PM restriction: monotonicity of the idealized dimensions (Proposition~\ref{prop:dimension-relationships}) and a positive-margin uniform-convergence motivation (Theorem~\ref{thm:cpvc-generalization}). Theorem~\ref{thm:cpvc-generalization} does not certify the empirical score; it identifies the region in which calibration-constrained category-level observations are compatible with standard margin-based uniform-convergence reasoning, assuming an external capacity envelope on the induced selection and residual classes. Neither result asserts that empirical category-level counts recover a true VC dimension. Full proofs and a positive-margin interpretation of the PVC proxy are in Appendix~\ref{app:proof-theorem-1}~and~\ref{appendix:sample_complexity}.

\begin{proposition}[Monotonicity of idealized dimensions]\label{prop:dimension-relationships}
For $\mathcal{F}$ containing deterministic predictors:
\textup{(i)} for $\gamma_1>\gamma_2$ in $(0,1]$, $\mathrm{VC}(\mathcal{F})=\mathrm{PVC}_1(\mathcal{F})\le\mathrm{PVC}_{\gamma_1}(\mathcal{F})\le\mathrm{PVC}_{\gamma_2}(\mathcal{F})$;
\textup{(ii)} for fixed $\gamma$ and $\tau_1>\tau_2$ in $[0,1)$, $\mathrm{C\text{-}PVC}_\gamma^{\tau_2}(\mathcal{F})\le\mathrm{C\text{-}PVC}_\gamma^{\tau_1}(\mathcal{F})\le\mathrm{PVC}_\gamma(\mathcal{F})$.
\end{proposition}

\noindent Classical VC equals $\mathrm{PVC}_1$; relaxing $\gamma$ monotonically increases capacity; calibration constraints can only reduce it. The PVC--C-PVC gap quantifies the empirical cost of calibration.

\paragraph{Positive-margin uniform convergence.}
For each evaluation example $z=(C,q,J(q))$, let $s_f(z)=\mathbf{1}\{\mathsf{Select}_f(q)=J(q)\}$ and $r_{f,\hat p}(z)=\hat p_f(q)-s_f(z)$, with induced classes $\mathcal{S}=\{s_f\}$ and $\mathcal{R}=\{r_{f,\hat p}\}$.

\begin{theorem}[Why the positive-margin region is statistically interpretable under capacity envelopes]\label{thm:cpvc-generalization}
Fix $(\gamma,\tau)$ with $\alpha=\gamma-\tau-\tfrac{1}{2}>0$, and assume the induced classes admit the capacity envelope $\mathrm{VC}(\mathcal{S})\le d_s$ and $\mathrm{fat}_\alpha(\mathcal{R})\le d_r$. Let $D_{\gamma,\tau}=d_s+d_r$. For any $\epsilon,\delta\in(0,1)$, if $m\ge\tilde O((D_{\gamma,\tau}+\log(1/\delta))/\epsilon^2)$, then with probability $\ge 1-\delta$,
\[
\sup_{s\in\mathcal{S}}\!\bigl|\tfrac{1}{m}\textstyle\sum_i s(z_i)-\mathbb{E}[s]\bigr|\le\epsilon
\;\;\text{and}\;\;
\sup_{r\in\mathcal{R}}\!\bigl|\tfrac{1}{m}\textstyle\sum_i r(z_i)-\mathbb{E}[r]\bigr|\le\epsilon.
\]
\end{theorem}

The proof (Appendix~\ref{app:proof-theorem-1}) combines VC uniform convergence on $\mathcal{S}$~\citep{anthony2009neural} with fat-shattering uniform convergence on $\mathcal{R}$~\citep{bartlett1995more, colomboni2025improved}. Crucially, $d_s, d_r$ are \emph{assumed complexity envelopes on the induced classes}, not the empirical C-PVC count; the theorem identifies the region (PM: $\gamma>\tau+1/2$) in which such an envelope can yield uniform convergence. The proxy reports what is observed; the theorem says where observations are theoretically interpretable.

\section{Experiments and Results}\label{sec:experiments}

\textbf{Released artifact.} The evaluation package is a \emph{complete evaluation suite}, not a new dataset: Math-360 testbed ($360$ problems, $8$ domains; full release), cross-domain splits for TruthfulQA / CSQA with verbatim prompts, judge-reference labels, model outputs for all evaluated models, the metric code for PVC / C-PVC / VUS / PM-VUS / SEA / CalibError, and scripts that reproduce every main-paper table and figure. The full artifact card is in Appendix~\ref{appendix:artifact_card} (Table~\ref{tab:artifact_card}). Three reproducibility levels are supported: \textbf{L1} main-paper tables from released outputs (no API; CPU, $<$10\,min); \textbf{L2} judge labels from released candidates (judge-API); \textbf{L3} end-to-end self-evaluation rerun (target-model access).

\subsection{Experimental Setup}

\textbf{Testbeds.} Math-360 is a controlled diagnostic testbed of $360$ original math problems across $8$ domains (balanced difficulty and subcategory), not a standalone general-purpose benchmark. Cross-domain validation uses TruthfulQA~\citep{lin-etal-2022-truthfulqa}, CommonsenseQA~\citep{talmor-etal-2019-commonsenseqa}, and MATH-500~\citep{huggingface2024math500}, each grouped into $10$ semantic categories with $240$ questions (Appendices~\ref{sec:math360}, \ref{sec:other_dataset}, \ref{appendix:math500}).

\textbf{Models.} Thirteen models: eleven 7--8B open models (SFT, SFT+RL, RL-only, distillation, DPO), Qwen2.5-32B-Instruct~\citep{yang2024qwen2}, and Claude 3.5 Haiku~\citep{anthropic2024haiku} as a proprietary mid-scale reference (decoding config in Appendix~\ref{sec:hyperparameters}).

\textbf{Reference labels.} Majority vote of three LLM judges (Claude 3.7 Sonnet, Amazon Nova Premier, DeepSeek-R1), invalid responses ($<0.4\%$) excluded, 3-0 agreement 53--75\% (Appendix~\ref{sec:judge_correlation}). The object of evaluation is \emph{reference-agreement self-evaluation}, not absolute correctness: all compared models face the same randomized majority-vote reference, and the diagnostic signal is the \emph{difference} between models under that fixed reference (risk mitigations: Appendix~\ref{appendix:judge_risks}). We report full-region VUS and PM-VUS ($\gamma > \tau + 1/2$); near-zero PM-C-PVC-VUS reads like a stress-test failure, not a metric defect: evidence that the model rarely enters a regime where high self-selection accuracy and tight calibration coexist. An exploratory SC-VUS proxy (not used in any main claim) is deferred to Appendix~\ref{appendix:sc-interpretation}.

\subsection{Diagnostic Decomposition Across Models}
\label{subsec:diagnostic_decomposition}

Table~\ref{tab:model_analysis_multi} is a decomposition, not a leaderboard. Full-region columns summarize category-level behavior across all $(\gamma,\tau)$; the PM-region column asks the stricter question of whether behavior survives under simultaneous accuracy and calibration constraints with a positive margin.

\begin{table}[t]
\centering
\small
\setlength{\tabcolsep}{5pt}
\renewcommand{\arraystretch}{1.15}
\resizebox{\linewidth}{!}{
\begin{tabular}{@{}l@{\hspace{6pt}}ccc@{\hspace{8pt}}c@{\hspace{8pt}}c@{\hspace{8pt}}c@{}}
\toprule
& \multicolumn{3}{c}{\cellcolor{blue!6}\textbf{Full Region}}
& \cellcolor{orange!6}\textbf{PM Region} ($\gamma>\tau+0.5$)
& \cellcolor{green!6}\textbf{Calibration}
& \cellcolor{purple!6}\textbf{Accuracy} \\
\cmidrule(lr){2-4} \cmidrule(lr){5-5} \cmidrule(lr){6-6} \cmidrule(lr){7-7}
\textbf{Model}
& \textbf{PVC-VUS} $\uparrow$ & \textbf{C-PVC-VUS} $\uparrow$ & \textbf{Gap} $\downarrow$
& \textbf{PM-C-PVC-VUS} $\uparrow$
& \textbf{CalibError} $\downarrow$
& \textbf{SEA} $\uparrow$ \\
\midrule
Qwen2.5-7B (Pretrain) \citep{yang2024qwen2}            & 5.26 & 3.76 & \underline{1.50} & 0.06 & 0.313 & 0.556 \\
Qwen2.5-7B-Instruct (SFT+RL) \citep{yang2024qwen2}     & \underline{5.61} & \underline{4.08} & 1.53 & \textbf{0.17} & 0.291 & \underline{0.599} \\
Qwen2.5-Math-7B-Instruct \citep{yang2024qwen25mathtechnicalreportmathematical} & 4.70 & 3.20 & \underline{1.50} & 0.002 & 0.335 & 0.506 \\
Llama-3.1-8B-Instruct (SFT+DPO) \citep{meta2024llama}  & 4.55 & 2.92 & 1.63 & 0.00 & 0.385 & 0.495 \\
OpenThinker2-7B (SFT) \citep{openthoughts2023openthinker} & 5.37 & 3.50 & 1.87 & 0.002 & 0.357 & 0.580 \\
DeepSeek-R1-Distill-Qwen-7B \citep{guo2025deepseek}    & 5.10 & 3.49 & 1.61 & 0.07 & 0.339 & 0.549 \\
Bespoke-Stratos-7B (SFT) \citep{bespokelabs2023stratos} & 5.33 & 3.54 & 1.79 & 0.00 & 0.347 & 0.570 \\
JiuZhang3.0-7B (SFT) \citep{zhou2024jiuzhang}          & 4.79 & 3.84 & \textbf{0.95} & 0.009 & \textbf{0.209} & 0.516 \\
Ministral-8B-Instruct-2410 \citep{mistralai2024ministral} & 4.88 & 3.00 & 1.88 & 0.00 & 0.402 & 0.522 \\
Open-Reasoner-Zero-7B (RL) \citep{hu2025openreasonerzeroopensourceapproach} & 5.29 & 3.51 & 1.78 & 0.02 & 0.352 & 0.566 \\
s1.1-7B (SFT) \citep{muennighoff2025s1simpletesttimescaling} & \textbf{5.83} & \textbf{4.27} & 1.56 & \underline{0.12} & \underline{0.285} & \textbf{0.622} \\
\bottomrule
\end{tabular}
}
\caption{\textbf{Diagnostic decomposition of LLM self-evaluation behavior (cross-dataset aggregate).} PVC-VUS summarizes category-level discrimination; the PVC--C-PVC Gap quantifies the empirical calibration cost; PM-C-PVC-VUS reports calibration-constrained behavior surviving in the positive-margin region $\gamma>\tau+0.5$ (Theorem~\ref{thm:cpvc-generalization}). \textbf{Bold}: best; \underline{underline}: second best. Per-dataset breakdowns (Math-360 / TruthfulQA / CSQA) and the exploratory PM-SC-VUS proxy are in Appendix~\ref{appendix:per_dataset_table}; MATH-500 results are in Appendix~\ref{appendix:math500}.}
\label{tab:model_analysis_multi}
\end{table}

Three observations emerge.

\textbf{(1) The expressivity--calibration separation is pervasive.}
Models with higher PVC-VUS (s1.1-7B $5.83$, Qwen2.5-7B-Instruct $5.61$) exhibit larger PVC--C-PVC gaps ($1.56$, $1.53$); JiuZhang3.0-7B has the smallest gap ($0.95$) at lower PVC-VUS ($4.79$), yet still has near-zero PM-C-PVC-VUS ($0.009$). No 7--8B model simultaneously achieves high PVC-VUS and a small gap.

\textbf{(2) Training paradigms reshape rather than close the tension.}
Instruction tuning, RLHF, distillation, and DPO improve raw capability, yet none yields nontrivial PM-C-PVC-VUS at 7--8B (max $0.17$, Qwen2.5-7B-Instruct). Higher PVC-VUS correlates with larger gaps: calibration cost grows precisely for the most discriminative models.

\textbf{(3) Scale helps discrimination but not the PM score (Table~\ref{tab:scale_and_ranking}).}
With Qwen2.5-32B-Instruct and Claude 3.5 Haiku, PVC-VUS improves modestly~\citep{kadavath2022language} but PM-C-PVC-VUS remains limited (Claude 3.5 Haiku $0.036$--$0.123$; Qwen2.5-32B-Instruct $0.148$--$0.209$ across the two pre-specified aggregation subsets). Ordering is preserved; the scarcity of PM-region behavior is not a 7--8B artifact.

\subsection{Aggregate UQ Can Reverse the Deployment Ranking}
\label{subsec:ranking_reversal}

\textbf{What other UQ metrics miss.}
AUROC/AUPRC test rank separability but are invariant to monotone transforms of confidence and ignore calibration. ECE/Brier test average confidence--accuracy alignment but admit cancellation across categories: high-confidence errors can be offset by low-confidence correct predictions. Selective-risk curves test a chosen operating point. None asks whether \emph{each} semantic category simultaneously satisfies accuracy and calibration constraints in the positive-margin region. Our diagnostics expose exactly that, and are complementary to existing UQ tools~\citep{lmpolygraph, uqlm, benchmark_uq}, not a replacement.

Table~\ref{tab:scale_and_ranking} combines the scale extension and the ranking-reversal comparison. Across the scale tiers, raw discrimination (PVC-VUS) improves modestly with scale; PM-C-PVC-VUS remains limited and the ordering is preserved. The aggregate-vs-PM split is the key signal: JiuZhang3.0-7B is the best model by ECE ($0.199$) and Brier ($0.049$) yet has PM-C-PVC-VUS only $0.019$, while Qwen2.5-32B-Instruct has worse ECE ($0.260$) but $\approx 11\times$ higher PM-C-PVC-VUS.

\begin{table}[t]
\centering
\small
\setlength{\tabcolsep}{5pt}
\renewcommand{\arraystretch}{1.15}
\resizebox{\linewidth}{!}{
\begin{tabular}{@{}llcccccc@{}}
\toprule
& & \multicolumn{2}{c}{\cellcolor{green!6}\textbf{Aggregate UQ}}
& \multicolumn{3}{c}{\cellcolor{blue!6}\textbf{Full Region}}
& \cellcolor{orange!6}\textbf{PM Region} \\
\cmidrule(lr){3-4} \cmidrule(lr){5-7} \cmidrule(lr){8-8}
\textbf{Model} & \textbf{Scale} & \textbf{ECE} $\downarrow$ & \textbf{Brier} $\downarrow$ & \textbf{PVC-VUS} $\uparrow$ & \textbf{C-PVC-VUS} $\uparrow$ & \textbf{Gap} $\downarrow$ & \textbf{PM-C-PVC-VUS} $\uparrow$ \\
\midrule
JiuZhang3.0-7B       & 7B          & \textbf{0.199} & \textbf{0.049} & 4.79 & 3.84 & \textbf{0.95} & 0.019 \\
Qwen2.5-7B-Instruct  & 7B          & 0.291 & 0.081 & 5.19 & 3.70 & 1.49 & 0.126 \\
Llama-3.1-8B-Instruct & 8B         & 0.385 & 0.140 & 4.40 & 2.86 & 1.55 & 0.000 \\
s1.1-7B              & 7B          & 0.293 & 0.094 & 5.44 & 3.94 & 1.50 & 0.142 \\
Qwen2.5-32B-Instruct & 32B         & 0.260 & 0.073 & 5.57 & 4.20 & 1.37 & \textbf{0.209} \\
Claude 3.5 Haiku     & Proprietary & 0.272 & 0.079 & 5.79 & 4.24 & 1.55 & 0.123 \\
\bottomrule
\end{tabular}}
\caption{Scale extension and ranking reversal combined. The aggregate-UQ winner (JiuZhang3.0-7B by ECE/Brier) is near-zero in PM-C-PVC-VUS, while Qwen2.5-32B-Instruct wins on PM despite worse aggregate metrics. PM-C-PVC-VUS here uses the ranking-reversal aggregation across the six models shown (same reference labels and $(\gamma,\tau)$ grid as Table~\ref{tab:model_analysis_multi}); Table~\ref{tab:model_analysis_multi} reports the broader cross-dataset aggregate over eleven 7--8B open models, so absolute values differ but the qualitative ordering is preserved. \textbf{Bold}: best.}
\label{tab:scale_and_ranking}
\end{table}

\textbf{Mechanism: bimodal confidence cancellation.}
Inspecting JiuZhang3.0-7B's confidence distribution explains the mechanism. Roughly $53.9\%$ of its predictions cluster near chance-level confidence ($\hat{p}\approx 0.50$, SEA $=0.494$), and $44.3\%$ concentrate at near-certainty ($\hat{p}\approx 0.985$, SEA $=0.583$: a $41.7\%$ error rate at near-full confidence). Averaging across these two regimes pulls aggregate calibration error down, while the high-confidence mass carries catastrophic miscalibration. Aggregate summaries observe only the average and report ``well calibrated''; the per-category C-PVC diagnostic shows that \emph{no} category simultaneously satisfies high accuracy and tight calibration in the PM region. This is not a cosmetic reformulation of existing summaries but a distinct deployment-relevant failure mode: for a confidence-gated pipeline, ECE/Brier recommend JiuZhang3.0-7B while Qwen2.5-32B-Instruct has $\approx 11\times$ higher PM-C-PVC-VUS.

\subsection{Confidence-Correctness Decoupling}
\label{subsec:conf_decoupling}

A stylistic-cue account would predict that reported confidence meaningfully separates correct from incorrect self-evaluations. The data do not support this. Across all $13$ models and $4$ datasets ($17{,}420$ predictions), the mean confidence gap between correct and incorrect decisions is only $0.005$. Across the $52$ model--dataset pairs, $21$ ($40\%$) show a \emph{negative} gap: models are on average more confident when they are wrong. The pattern is consistent per domain: Math-360 $0.004$, MATH-500 $0.007$, CSQA $0.005$, TruthfulQA $0.000$. Restricting to unanimous-judge pairs (cleaner reference), the gap remains only $0.008$ with a $38.0\%$ high-confidence error rate at confidence $\ge 0.9$. Discrimination is partially real, but confidence is not actionable as a deployment gate.

\subsection{Robustness}
\label{subsec:robustness}

\textbf{Category robustness.}
Rankings are stable under $7$ alternative categorizations (random partitions (3 seeds), difficulty-based splits, and subcategory variants) across all $4$ datasets, with C-PVC-VUS Spearman rank correlation in $[0.75, 0.92]$. This rules out the concern that the ordering is an artifact of the particular semantic partitioning.

\textbf{Judge-label perturbation.}
Synthetically flipping $10$--$30\%$ of judge labels (mean $\pm$ one s.d.\ over $5$ seeds per level) degrades PVC-VUS gradually, as expected from standard concentration. The near-zero PM-C-PVC-VUS regime for models already at that level is stable: JiuZhang3.0-7B stays at $0.019$--$0.023$ under $30\%$ corruption. A label-noise explanation would either depress high-PM models toward zero or inflate low-PM models; neither happens. The central finding is therefore not a judge-noise artifact (Appendix~\ref{appendix:label-perturbation}).

\textbf{Unanimous-only re-analysis.}
Restricting to samples with 3-0 judge agreement raises mean PM-C-PVC-VUS from $0.073$ to $0.320$ while preserving the ranking structure (JiuZhang remains near zero at $0.047$; s1.1-7B rises to $0.996$). The cleaner-label condition does not erase the main distinction (Appendix~\ref{appendix:unanimous}).

\subsection{Visual Signatures and Cross-Domain Analysis}

\begin{figure}[!t]
\centering
\begin{subfigure}[b]{0.48\linewidth}
    \centering
    \includegraphics[width=\linewidth]{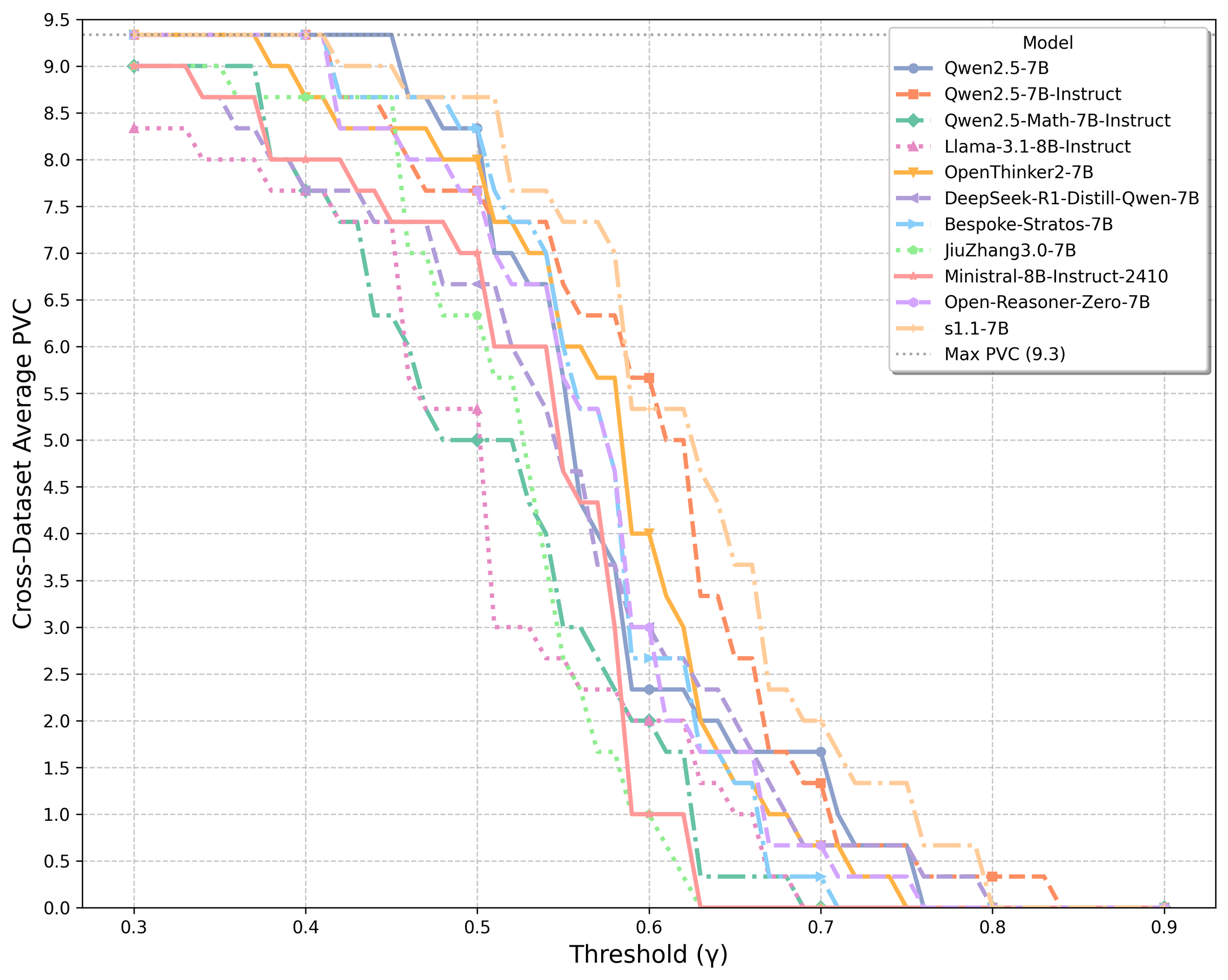}
    \caption{PVC across $\gamma$.}
    \label{fig:cross_dataset_average_pvc_plot}
\end{subfigure}\hfill
\begin{subfigure}[b]{0.48\linewidth}
    \centering
    \includegraphics[width=\linewidth]{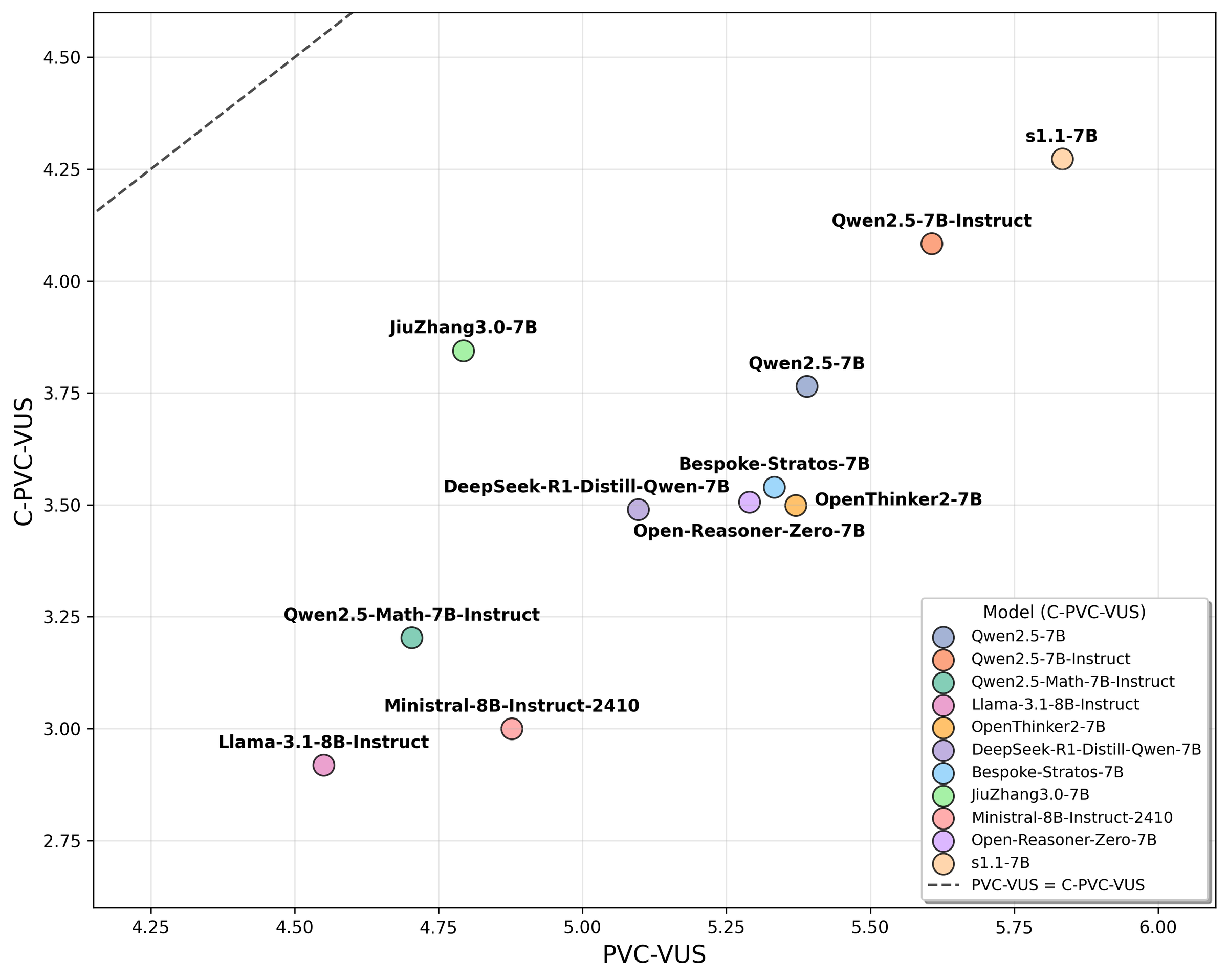}
    \caption{PVC-VUS vs.\ C-PVC-VUS.}
    \label{fig:cross_domain_pvc_cpvc_comparison}
\end{subfigure}
\caption{(a) All models plateau near maximum PVC at lenient $\gamma$ but decline to zero near $\gamma=1$. (b) Scatter of PVC-VUS vs.\ C-PVC-VUS; the diagonal is perfect calibration. JiuZhang3.0-7B lies closest; s1.1-7B achieves highest PVC-VUS at greater calibration cost.}
\label{fig:combined_calibration_pvc_plot}
\end{figure}

\begin{figure}[!t]
    \centering
    \includegraphics[width=0.95\linewidth]{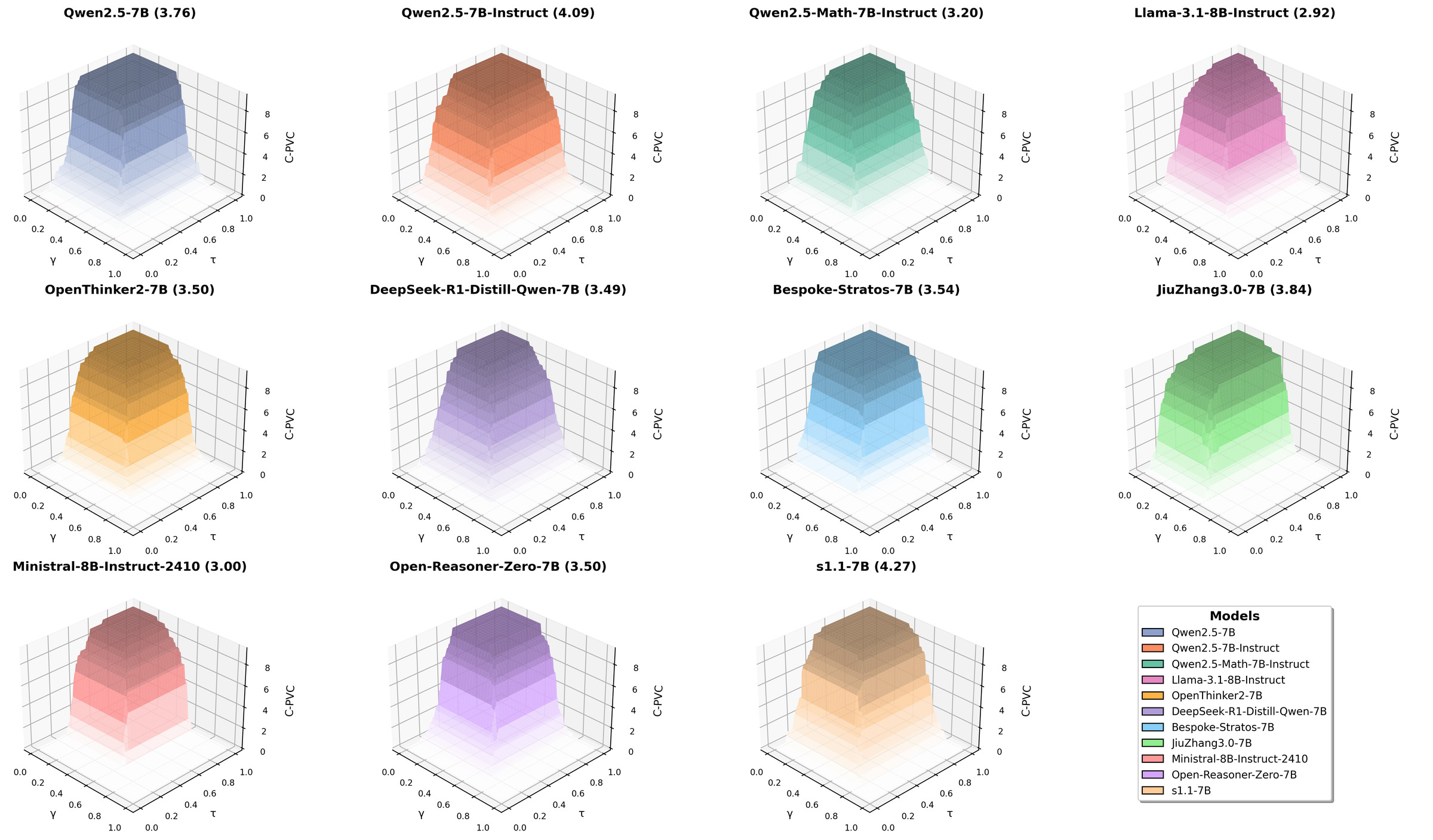}
    \caption{Empirical C-PVC diagnostic surfaces for eleven 7--8B models ($\tau$: calibration tolerance, $\gamma$: accuracy threshold, $z$: empirical C-PVC proxy). Distinct topographical profiles give per-model signatures; C-PVC-VUS (parentheses) aggregates each surface. Extended analysis in Appendix~\ref{appendix:cross_domain_vis}.}
    \label{fig:combined_model_performance}
\end{figure}

Figure~\ref{fig:combined_calibration_pvc_plot} summarizes the threshold--calibration behavior across models; Figure~\ref{fig:combined_model_performance} shows the full $3$D C-PVC surfaces. Domain-specific signatures: OpenThinker2-7B peaks on Math-360, Qwen2.5-7B-Instruct on TruthfulQA, s1.1-7B on CSQA; JiuZhang3.0-7B has the smallest PVC--C-PVC gap across all three.

\section{Related Work}

\textbf{LLM reasoning and self-reflection.} Large language models have achieved strong performance on complex reasoning tasks \citep{kojima2022large, wei2022chain, chen2022program}, evaluated on benchmarks including MathVista \citep{li2024understanding}, Bongard-Logo \citep{nie2020bongard}, RAVEN \citep{zhang2019raven}, and ARC \citep{chollet2019measure}. Techniques such as DPO \citep{rafailov2023direct} and GRPO \citep{shao2024deepseekmath, guo2025deepseek, yang2024qwen2, yu2025dapo} further enhance capabilities. Self-reflection research spans educational psychology \citep{brown_visualizing_2010, govaerts_student_2012, johnston_amplifying_2005, leijen_streaming_2009, nelson_push-me_2008, palinscar_reciprocal_1984, reznitskaya_examining_2012, tseng_design_2013, webb_promoting_2013} to LLM methodologies \citep{shi_enhancing_2019, bentvelzen2022revisiting}. Contemporary approaches (Reflexion \citep{shinn2023reflexion}, Self-Refine \citep{madaan2023self}, CRITIC \citep{pan2023automatically}, and the iterative DPO/SFT R$^3$ pipeline~\citep{wang2026reflect}) implement iterative refinement (typically with an external larger-model teacher that a calibrated self-evaluator could in principle replace), while Metacognitive Prompting \citep{toy2024metacognition} and metacognitive language models \citep{wang2023metacognitive} attempt formalization. However, most frameworks prioritize final accuracy over assessing reasoning quality \citep{mondorf2024beyond}.

\textbf{Calibration and uncertainty quantification.} Calibration has been studied extensively~\citep{guo2017calibration, ovadia2019can, kadavath2022language}, with techniques including temperature scaling~\citep{platt1999probabilistic}, conformal prediction~\citep{angelopoulos2021gentle}, deep ensembles~\citep{lakshminarayanan2016simple}, Bayesian approximations~\citep{gal2015dropout, woo2022analytic}, and balanced entropy measures~\citep{jaeoh_iclr}. Recent LLM-specific UQ benchmarks and tools include LM-Polygraph~\citep{lmpolygraph}, UQLM~\citep{uqlm}, and the broader benchmarking of LLM uncertainty~\citep{benchmark_uq}. Confidence elicitation from LLMs has also been studied empirically~\citep{xiong2024canllms}. Our framework is complementary to these tools rather than a replacement: aggregate UQ scores (ECE, Brier, semantic-entropy, ensemble scorers) report a distribution-level summary and can admit cancellation effects; our per-category, calibration-constrained diagnostics instead ask whether calibration-constrained self-evaluation behavior survives uniformly within the PM region (Section~\ref{subsec:ranking_reversal}).

\textbf{LLM-as-a-judge: bias, position, and scale effects.} Model-based judging has well-documented limitations: LLM judges can favor their own generations~\citep{panickssery2024llmjudge}, exhibit position and length biases~\citep{shi2025position}, and only the largest judges reliably align with humans~\citep{judgingthejudges2025}. We partially mitigate these effects via a three-judge majority vote with randomized solution order and report agreement statistics (Appendix~\ref{sec:judge_correlation}) and label-perturbation robustness (Section~\ref{subsec:robustness}); we do not claim judge labels are oracles.

\textbf{Alignment and optimization.} RLHF \citep{ouyang2022training}, DPO \citep{rafailov2023direct}, Contrastive Preference Learning \citep{hejna2023contrastive}, self-rewarding mechanisms \citep{yuan2024self}, and GRPO \citep{shao2024deepseekmath} have enhanced LLM capabilities \citep{swamy2025all}. Our C-PVC metric evaluates whether such training improves introspective reliability or merely suppresses calibration errors.

\textbf{Evaluation design and protocol choice.} A growing body of work documents that evaluation design can change conclusions: holistic multi-scenario benchmarks make method comparisons more robust~\citep{liang2023helm}; large collaborative benchmarks show that single-metric leaderboards understate capability variation~\citep{srivastava2023bigbench}; and behavioral audits show that task-frequency and prompt phrasing systematically bias apparent performance~\citep{mccoy2024embers}. Our diagnostics are in the same spirit, applied to the specific setting of self-reported confidence: we show that the \emph{aggregation rule} of a calibration metric (full-surface vs.\ positive-margin, per-category vs.\ pooled) can reverse deployment-level conclusions under fixed reference labels and fixed models.

\textbf{Theoretical foundations.} The VC dimension \citep{vapnik1971theory, blumer1989learnability, shalev2014understanding} and fat-shattering dimension \citep{alon1997scale, bartlett1995more, bartlett1994fat} provide foundations for model capacity and generalization. Extensions to probabilistic predictors \citep{klesk2012probabilistic} and neural networks \citep{mendelson2003entropy, anthony2009neural, telgarsky2017neural, bartlett2019nearly, colomboni2025improved} have not been applied to LLM self-evaluation. Our PVC/C-PVC diagnostics are VC/fat-shattering motivated but are reported as empirical evaluation proxies rather than literal capacity estimates, providing a theory-motivated way to study when calibration-constrained self-evaluation survives in the PM region.

\section{Conclusion}

We proposed a black-box, calibration-constrained evaluation framework for LLM self-evaluation. The framework introduces two theory-motivated, operational diagnostics (PVC and C-PVC) and a positive-margin restriction PM-VUS, which is intentionally stringent: it asks whether calibrated self-evaluation survives in the positive-margin region where Theorem~\ref{thm:cpvc-generalization} provides a uniform-convergence motivation under an assumed capacity envelope.

Across 13 models from 7B to 32B plus a proprietary mid-scale system and four benchmarks, we identify a deployment-relevant ranking reversal: the ECE/Brier-best model (JiuZhang3.0-7B) has near-zero PM-C-PVC-VUS while Qwen2.5-32B-Instruct (worse ECE) has $\approx 11\times$ higher PM-C-PVC-VUS. The mechanism is bimodal confidence clustering that cancels in averages but fails per-category calibration. Complementary analyses (mean correct--incorrect gap $0.005$ across $17{,}420$ predictions, categorization robustness at Spearman $\ge 0.75$, stability under $10$--$30\%$ judge-label perturbation) rule out categorization, judge noise, and model scale as artifacts. Optimizing directly for PM-C-PVC-VUS and extending to open-ended or multi-turn self-evaluation are natural next steps; claim boundaries and limitations are in Appendices~\ref{appendix:limitations}--\ref{appendix:claim_boundary}.

\newpage

\bibliographystyle{plainnat}
\small
\bibliography{references}

\clearpage
\appendix

\section{Artifact / Evaluation Card}
\label{appendix:artifact_card}

This appendix summarizes the released evaluation artifacts in a Dataset/Evaluation Card format, intended to make intended use, limitations, provenance, and reproducibility explicit for downstream users.

\begin{table}[!ht]
\centering
\footnotesize
\renewcommand{\arraystretch}{1.1}
\setlength{\tabcolsep}{5pt}
\begin{tabular}{@{}p{0.22\linewidth} p{0.73\linewidth}@{}}
\toprule
\textbf{Field} & \textbf{Description} \\
\midrule
Artifact name & LLM Self-Evaluation Diagnostics Package (v1.0): protocol, metrics, testbed, and scripts. \\
Components &
(i) three-stage self-evaluation protocol (two candidate-generation prompts, self-selection prompt, judge prompt, under \texttt{prompts/});
(ii) Math-360 (\texttt{data/math360/}; $360$ original problems, $8$ domains, $5$ subcategories, difficulty-stratified);
(iii) cross-domain category splits for TruthfulQA, CommonsenseQA, MATH-500 (\texttt{data/cross\_domain/});
(iv) model outputs (\texttt{model\_outputs/*\_self\_eval.csv});
(v) three-judge reference labels (\texttt{reference\_labels/}; schema documented);
(vi) \texttt{selfeval} package (models, judges, evaluation, experiment, utils) with YAML configs;
(vii) scripts for PVC/C-PVC computation, cross-dataset aggregation, end-to-end evaluation, main-table reproduction, and reference-label rebuild;
(viii) documentation (README, reviewer README, release-strategy, dataset/evaluation cards, Croissant metadata). \\
\midrule
Intended use & Diagnosing reliability of LLM self-reported confidence under per-category accuracy and calibration constraints; comparing models under a fixed reference signal; ablating evaluation assumptions. \\
Not intended use & General-purpose math leaderboard; absolute human-level capability ranking; single-scalar replacement for ECE/Brier/AUROC; any claim that a model's self-evaluation is reliable outside the declared protocol. \\
\midrule
Math-360 source & Original problems authored for this work, stratified across $8$ domains (Arithmetic, Algebra, Calculus, Geometry, Number Theory, Combinatorics, Statistics, Linear Algebra), $5$ subcategories each, difficulty-stratified (Easy / Medium / Hard). Released in full. \\
Reused datasets & TruthfulQA~\citep{lin-etal-2022-truthfulqa}, CommonsenseQA~\citep{talmor-etal-2019-commonsenseqa}, MATH-500~\citep{huggingface2024math500}. Question text is \emph{not} redistributed; we release only IDs, category mappings, and derived outputs. Sampling in Appendix~\ref{sec:other_dataset}. \\
\midrule
Reference-labeling & Three-judge LLM ensemble (Claude 3.7 Sonnet, Amazon Nova Premier, DeepSeek-R1); pairwise binary preference; majority vote; invalid responses ($<0.4\%$) excluded. See \S\ref{subsec:measuring_pvc_cpvc} and Table~\ref{tab:judge_risks}. \\
Reference is not & A human-validated ground truth; an oracle of mathematical correctness; a guarantee of absolute correctness for any individual pair. \\
\midrule
Known limitations & Judge bias / self-preference (mitigated via three-judge majority with randomized order); category dependence (stable under $7$ alternative categorizations, Spearman $\ge 0.75$); small controlled math set of $360$ items (validated by cross-domain replication); proprietary judge versions may drift (pinned at experiment time). \\
\midrule
Version / frozen split & Math-360 v1.0; TruthfulQA/CSQA/MATH-500 splits v1.0 (Appendix~\ref{sec:other_dataset}). Future revisions will be versioned. \\
License & Math-360 original data, reference labels, and derived outputs: \textbf{CC-BY-4.0}. \texttt{selfeval} code and reproduction scripts: \textbf{Apache-2.0}. Reused datasets retain upstream licenses. \\
Hosting & Code and data are hosted together at the public GitHub repository \url{https://github.com/calibration-constrained-diagnostics/selfeval-diagnostics/}. Croissant metadata accompanies the data release. \\
\midrule
Reproducibility & Three levels: \textbf{L1} metric-only reproduction of every main-paper table from released \texttt{model\_outputs/} ($<$10\,min, CPU); \textbf{L2} judge-label recomputation with judge-API access; \textbf{L3} end-to-end self-evaluation reruns with target-model access. Prompts, configs, and $(\gamma,\tau)$ grid are pinned in the release. \\
\bottomrule
\end{tabular}
\caption{Artifact / evaluation card for the released diagnostic package.}
\label{tab:artifact_card}
\end{table}

\section{Use of Large Language Models in Manuscript Preparation}
\label{appendix:llm_usage}

The authors used LLM-based assistants during the preparation of this manuscript. This use was limited to editorial support: grammar and wording suggestions, improving the readability of author-written text, and checking \LaTeX{} formatting. Text produced with such assistance was treated as a draft only; the authors reviewed, revised, and verified every passage before including it.

The research questions, the evaluation protocol, the PVC/C-PVC definitions, the theoretical statements and proofs, the experimental design, and the interpretation of the results are the work of the authors. LLM assistants were not used to produce or modify reported results. LLMs are not authors of this paper. The authors take full responsibility for the entire content of the paper, including any errors that may remain.

This appendix covers LLM use \emph{for manuscript preparation only}. The separate, methodologically declared use of LLMs as (i) the subject of evaluation and (ii) a reference-labeling ensemble is described in \S\ref{subsec:measuring_pvc_cpvc}, \S\ref{sec:experiments}, Appendix~\ref{sec:judge_correlation}, and Appendix~\ref{sec:detailed-experimental-methodology}.

\section{Limitations and Broader Impacts}
\label{appendix:limitations}

\paragraph{Limitations.}
Several limitations apply. (i) The main table targets 7--8B open models; the scale extension adds Qwen2.5-32B-Instruct and Claude 3.5 Haiku but not frontier-scale systems. (ii) Empirical PVC/C-PVC are operational diagnostic proxies under black-box access, not literal VC-dimension estimators; exact shattering is unverifiable and the true hypothesis class is unobservable. (iii) Reference labels come from a three-judge LLM ensemble rather than human annotation; majority voting with label-perturbation robustness (Appendix~\ref{appendix:label-perturbation}) and unanimous-only re-analysis (Appendix~\ref{appendix:unanimous}) support comparative use, but some label noise persists and proprietary judge versions may drift. (iv) The two-solution pairwise protocol is a specific reduction of self-evaluation; open-ended generation and multi-turn self-critique are not directly covered. (v) Tasks are reasoning-centric (Math-360, MATH-500, TruthfulQA, CSQA); behavior on subjective or creative tasks is future work. (vi) The framework is diagnostic and does not propose a training remedy.

\paragraph{Broader impacts.}
This paper is primarily diagnostic: it releases tools and benchmark procedures for measuring existing model behavior rather than introducing new model capabilities. The main intended positive impact is helping practitioners identify models prone to overconfident yet incorrect self-assessments before deployment, reducing the risk of silent failures in confidence-gated autonomous pipelines. The main residual risk is misuse of the diagnostic as a false sense of security: favorable PVC/C-PVC scores in evaluated categories do not guarantee reliability in untested settings. We therefore emphasize that PM-C-PVC-VUS should be read alongside, not in place of, other uncertainty metrics and domain-specific safety evaluations. A detailed claim-boundary table (what the artifact enables and does not enable) is in Appendix~\ref{appendix:claim_boundary}.

\section{LLMs as Function Classes: Three Views}
\label{appendix:llm_function_class}

The category-level diagnostic framework treats an LLM as a \emph{family} of predictors rather than a single fixed function. Three well-founded perspectives motivate this view.

\textbf{Bayesian view (parameter-level randomization).} Treating the parameter vector $W$ as a random variable $W\sim\mathcal{P}$ induces $\{f_W:W\in\mathrm{supp}(\mathcal{P})\}$. This is standard in PAC-Bayesian theory~\citep{mcallester2003simplified, germain2016pac}, underlies the NN--GP correspondence~\citep{lee2017deep, matthews2018gaussian}, and has yielded non-vacuous generalization bounds~\citep{dziugaite2017computing}.

\textbf{Prompt-based view (input-conditioning).} With fixed parameters $W$, varying a prompt $p\in\Pi$ induces $\mathcal{F}_{\mathrm{prompt}}(W)=\{x\mapsto f_W([p;x]):p\in\Pi\}$, consistent with prompts functioning as task-specific adapters~\citep{li2021prefix, lester2021power, xie2022explanation}.

\textbf{Stochastic decoding view (output-level randomization).} Temperature/nucleus sampling induces a distribution over outputs, mapping the model to a probabilistic predictor aligned with Gibbs predictors~\citep{mcallester2003simplified} and MC-dropout Bayesian approximation~\citep{gal2015dropout}.

Combining these, we fix $W$ and vary prompts ($\Pi$) and decoding ($\Omega$): $\mathcal{F}(W)=\{f(\cdot;W,\pi,\omega):\pi\in\Pi,\omega\in\Omega\}$. This is the operational hypothesis class to which our idealized capacity notions are applied.

\section{Operational Positioning of the PVC/C-PVC Proxies}
\label{appendix:operational_proxy}

The empirical PVC/C-PVC quantities are \emph{tractable, theory-motivated diagnostics under black-box access}, not literal VC-dimension estimators. Direct verification of all $2^d$ labelings is intractable, and the true hypothesis class $\mathcal{F}$ is unobservable under black-box access. What the category-level count provides is not a certified lower bound on the true hypothesis-class dimension, but an \emph{operational diagnostic count under the chosen prompt/decoding protocol}. This positioning preserves the structural separation between discrimination, per-category calibration, and positive-margin region behavior, the three quantities that aggregate uncertainty summaries conflate.

\paragraph{Theory-to-practice mapping.} (i) inputs $x_i \mapsto$ semantic categories $C$; (ii) shattering condition $\mathbb{P}(f(x_i)=y_i^*)\ge\gamma \mapsto$ empirical $\mathrm{SEA}(f,C)\ge\gamma$; (iii) calibration condition $|\hat{p}(x_i)-\mathbb{E}[\mathbf{1}\{f(x_i)=y_i^*\}]|\le\tau \mapsto$ category-level $\mathrm{CalibError}(f,C)\le\tau$. Under this mapping, empirical PVC/C-PVC counts categories satisfying the corresponding operational constraints. These counts can be interpreted as lower-bound evidence for the idealized capacity notion \emph{only under explicit richness assumptions on the prompt/decoding-induced family}; we report them as operational evaluation proxies, not distribution-free estimates of the true hypothesis-class dimension. Categories are \emph{distributional evaluation units}, not literal VC inputs.

\section{Reference-Label Risks and Mitigations}
\label{appendix:judge_risks}

Table~\ref{tab:judge_risks} summarizes the reference-label risks we anticipate, the mitigations built into the protocol, and the limitations that remain after those mitigations.

\begin{table}[H]
\centering
\small
\renewcommand{\arraystretch}{1.15}
\setlength{\tabcolsep}{6pt}
\resizebox{\linewidth}{!}{
\begin{tabular}{@{}lll@{}}
\toprule
\textbf{Risk} & \textbf{Mitigation} & \textbf{Remaining limitation} \\
\midrule
Judge bias / self-preference & 3-judge ensemble, randomized presentation order & Not a human oracle \\
Label noise & $10$--$30\%$ synthetic label perturbation (5 seeds) & Synthetic corruption only \\
Ambiguous pairs (both correct / both wrong) & Unanimous-only (3-0) re-analysis & Reduces effective sample \\
Style-preference artifacts & Same reference applied across all compared models & May still reward judge-preferred style \\
\bottomrule
\end{tabular}}
\caption{Summary of reference-label risks, mitigations, and remaining limitations.}
\label{tab:judge_risks}
\end{table}

\section{Claim Boundary: What This Artifact Enables and Does Not Enable}
\label{appendix:claim_boundary}

Table~\ref{tab:claim_boundary} states the claims that the released diagnostic package is designed to support, alongside the claims that it is explicitly \emph{not} intended to support. This accompanies the per-field declarations in the artifact card (Appendix~\ref{appendix:artifact_card}).

\begin{table}[H]
\centering
\small
\renewcommand{\arraystretch}{1.15}
\setlength{\tabcolsep}{6pt}
\begin{tabular}{@{}p{0.47\linewidth} p{0.47\linewidth}@{}}
\toprule
\textbf{Claims this artifact enables} & \textbf{Claims this artifact does \emph{not} enable} \\
\midrule
Compare whether model-reported confidence tracks judge-referenced self-selection correctness across models. & Certify absolute correctness of model reasoning. \\
Detect category-level calibration/discrimination failures hidden by aggregate UQ summaries. & Provide human-preference alignment without additional human validation. \\
Analyze whether calibration-constrained self-evaluation behavior survives in the positive-margin (PM) region. & Literally estimate VC dimension or derive distribution-free sample-complexity bounds. \\
Reproduce and extend PVC/C-PVC/PM-VUS surfaces across new models and datasets under the declared protocol. & Make universal claims that LLMs cannot self-evaluate outside the declared protocol. \\
Diagnose deployment-relevant failure modes for confidence-gated downstream decisions. & Certify deployment safety of any specific model. \\
\bottomrule
\end{tabular}
\caption{Claim boundaries for the released evaluation package.}
\label{tab:claim_boundary}
\end{table}

\section{Notation}\label{appendix:notation}

Table~\ref{tab:notation} collects the notation used throughout the paper.

\begin{table}[H]
\centering
\small
\renewcommand{\arraystretch}{1.1}
\begin{tabular}{cl}
\toprule
\textbf{Symbol} & \textbf{Description} \\
\midrule
$\mathcal{X}$ & Input space (set of problem categories) \\
$\mathcal{F}$ & Class of probabilistic predictors $f: \mathcal{X} \rightarrow \Delta(\{0,1\})$ \\
$\Delta(\{0,1\})$ & Set of probability distributions over binary output space $\{0,1\}$ \\
$\gamma \in (0,1]$ & Accuracy threshold parameter \\
$\tau \in [0,1)$ & Calibration error tolerance parameter \\
$\mathrm{VC}(\mathcal{F})$ & Classical Vapnik-Chervonenkis dimension of function class $\mathcal{F}$ \\
$\mathrm{PVC}_\gamma(\mathcal{F})$ & Probabilistic VC dimension at accuracy threshold $\gamma$ \\
$\mathrm{C\text{-}PVC}_\gamma^\tau(\mathcal{F})$ & Calibration-aware PVC dimension with parameters $\gamma$ and $\tau$ \\
$\hat{p}: \mathcal{X} \to [0,1]$ & Confidence scoring function associated with predictor $f$ \\
$\mathrm{fat}_\alpha(\mathcal{F})$ & Fat-shattering dimension at scale $\alpha$ \\
$\alpha = \gamma - \frac{1}{2}$ & Margin parameter relating PVC to fat-shattering dimension \\
$\mathbb{P}(f(x) = y)$ & Probability that predictor $f$ outputs label $y$ on input $x$ \\
$\mathbb{E}[\cdot]$ & Expectation operator \\
$\mathbf{1}\{\cdot\}$ & Indicator function \\
$\mathrm{SEA}(f, C)$ & Self-Evaluation Accuracy of model $f$ on category $C$ \\
$\mathsf{Judge}(q)$ & Reference preference label (three-judge majority vote) for question $q$ \\
$\mathsf{Select}_f(q)$ & Model $f$'s selected solution index for question $q$ \\
$\mathrm{CalibError}(f, C)$ & Calibration error of model $f$ on category $C$ \\
$M\text{-VUS}(\mathcal{F})$ & Volume Under Surface for metric $M \in \{\mathrm{PVC}, \mathrm{C\text{-}PVC}, \mathrm{SC}\}$ \\
$\mathrm{SC}(\gamma, \tau, \mathcal{F})$ & Sample complexity bound with parameters $\gamma$, $\tau$ \\
$\epsilon$ & Generalization error tolerance \\
$\delta$ & Confidence parameter in generalization bounds \\
$K$ & Universal constant in sample complexity bounds \\
$\Pi$ & Set of prompts in prompt-based function class \\
$W$ & Parameter vector of a language model \\
$\mathcal{P}$ & Probability distribution over parameter space \\
$\mathcal{D}$ & Distribution over input space $\mathcal{X}$ \\
$\mathcal{D}_C$ & Distribution of questions within category $C$ \\
\bottomrule
\end{tabular}
\caption{Notation used throughout the paper.}
\label{tab:notation}
\end{table}

\section{Proof of Proposition \ref{prop:dimension-relationships}}

\begin{proof}
We prove the two statements in order.

\textbf{Statement 1.} For any $\gamma_1, \gamma_2 \in (0,1]$ with $\gamma_1 > \gamma_2$:
\[
\mathrm{VC}(\mathcal{F}) = \mathrm{PVC}_1(\mathcal{F}) \le \mathrm{PVC}_{\gamma_1}(\mathcal{F}) \le \mathrm{PVC}_{\gamma_2}(\mathcal{F}).
\]

By definition, $\mathrm{PVC}_\gamma(\mathcal{F})$ is the largest integer $d$ such that for every labeling $(y_1^*, \dots, y_d^*)$, there exists $f \in \mathcal{F}$ satisfying $\mathbb{P}(f(x_i) = y_i^*) \ge \gamma$ for all $i \in [d]$.

When $\gamma = 1$, this requires $\mathbb{P}(f(x_i) = y_i^*) = 1$. Under the assumption that $\mathcal{F}$ contains deterministic predictors, this coincides with classical VC shattering, yielding $\mathrm{PVC}_1(\mathcal{F}) = \mathrm{VC}(\mathcal{F})$.

For monotonicity, suppose $\gamma_1 > \gamma_2$. Any $\gamma_1$-shattered set is also $\gamma_2$-shattered since the weaker threshold is easier to satisfy. Thus $\mathrm{PVC}_{\gamma_1}(\mathcal{F}) \le \mathrm{PVC}_{\gamma_2}(\mathcal{F})$, completing the first chain.

\textbf{Statement 2.} For fixed $\gamma \in (0,1]$ and $\tau_1 > \tau_2$ in $[0,1)$:
\[
\mathrm{C\text{-}PVC}_\gamma^{\tau_2}(\mathcal{F}) \leq \mathrm{C\text{-}PVC}_\gamma^{\tau_1}(\mathcal{F}) \leq \mathrm{PVC}_\gamma(\mathcal{F}).
\]

First, $\mathrm{C\text{-}PVC}_\gamma^{\tau_1}(\mathcal{F}) \leq \mathrm{PVC}_\gamma(\mathcal{F})$ follows directly from definitions. C-PVC requires both accuracy ($\mathbb{P}(f(x_i) = y_i^*) \ge \gamma$) and calibration ($|\hat{p}(x_i) - \mathbb{E}[\mathbf{1}\{f(x_i) = y_i^*\}]| \le \tau$). The additional calibration constraint can only reduce or maintain the dimension.

Second, $\mathrm{C\text{-}PVC}_\gamma^{\tau_2}(\mathcal{F}) \leq \mathrm{C\text{-}PVC}_\gamma^{\tau_1}(\mathcal{F})$ holds because $\tau_2 < \tau_1$ imposes a stricter calibration requirement. Any predictor satisfying calibration tolerance $\tau_2$ automatically satisfies $\tau_1$, so any $\tau_2$-shattered set is also $\tau_1$-shattered.

Combining these inequalities completes the proof.
\end{proof}

\section{Positive-Margin Interpretation of the PVC Proxy}\label{appendix:sample_complexity}

The empirical PVC proxy reported in the main tables is not itself a sample-complexity parameter. It counts observed categories satisfying the $\gamma$-constraint and is a diagnostic indicator, not a distribution-free capacity estimate. The next proposition clarifies the role of PVC: it gives a \emph{lower-bound} relationship between PVC shattering and fat-shattering of an associated real-valued class at a positive margin, which is a \emph{margin interpretation} of the $\gamma$-constraint. This is not a generalization bound: lower bounds on class complexity do not control uniform convergence.

\begin{proposition}[PVC shattering implies positive-margin fat-shattering]
\label{thm:sample-complexity-pvc}
Let $\mathcal{F}$ be a class of probabilistic predictors $f: \mathcal{X} \to \Delta(\{0,1\})$, identified with pairs $(f_0,f_1)$ where $f_0(x) + f_1(x) = 1$. Fix $\gamma \in (1/2, 1]$. Let $\mathcal{G} = \{f_1 : f \in \mathcal{F}\} \subseteq [0,1]^\mathcal{X}$ and $\alpha = \gamma - 1/2 > 0$. Then (Lemma~\ref{lem:pvc-fat})
\[
\mathrm{PVC}_\gamma(\mathcal{F}) \le \mathrm{fat}_\alpha(\mathcal{G}).
\]
\end{proposition}

\paragraph{Reading of the proposition.} This inequality says that if the PVC constraint is satisfied by a set of size $d$, then $\mathcal{G}$ fat-shatters a set of size at least $d$ at margin $\alpha$. It is a structural \emph{lower bound} on the fat-shattering dimension of $\mathcal{G}$, not an upper bound. Consequently, it does \emph{not} imply a sample-complexity guarantee in terms of $\mathrm{PVC}_\gamma(\mathcal{F})$.

\paragraph{What would be needed for a formal bound.} A formal uniform-convergence bound for $\mathcal{G}$ requires an \emph{upper bound} $\mathrm{fat}_\alpha(\mathcal{G}) \le D$ for some capacity parameter $D$. When such a $D$ is available, the standard fat-shattering uniform-convergence inequality gives, for a universal constant $K$, $\sup_{g\in\mathcal{G}} |m^{-1}\sum_i g(x_i) - \mathbb{E}[g(x)]| \le \epsilon$ whenever $m \ge \tilde{O}((D + \log(1/\delta))/\epsilon^2)$. The PVC proxy does not estimate such a $D$; it is a diagnostic indicator.

\paragraph{Relation to the empirical proxy and SC-VUS.} The empirical PVC count in the main tables is an operational indicator of how many observed categories satisfy the $\gamma$-constraint, under the chosen prompt/decoding protocol. It is useful as a comparative diagnostic across models and thresholds, but it is not a distribution-free capacity estimate and not an absolute sample-size guarantee. This is why the main text reports SC-VUS as a \emph{relative verification-cost index} rather than an absolute sample-size bound.

\begin{lemma}[PVC shattering implies fat-shattering at a positive margin]\label{lem:pvc-fat}
Let $\mathcal{F}$ be a class of probabilistic predictors, identified with pairs $(f_0,f_1)$ with $f_0+f_1=1$, and let $\mathcal{G} = \{f_1 : f \in \mathcal{F}\} \subseteq [0,1]^{\mathcal{X}}$. Fix $\gamma \in (1/2,1]$ and set $\alpha = \gamma - 1/2 > 0$. Then
\[
\mathrm{PVC}_\gamma(\mathcal{F}) \le \mathrm{fat}_\alpha(\mathcal{G}),
\]
with fat-shattering evaluated at fixed thresholds $s_i = 1/2$.
\end{lemma}

\begin{proof}
Suppose $\{x_1, \dots, x_d\}$ is $\gamma$-shattered in the PVC sense. For any labeling $(y_1^*,\dots,y_d^*) \in \{0,1\}^d$, there exists $f = (f_0, f_1) \in \mathcal{F}$ such that $f_{y_i^*}(x_i) \ge \gamma$ for all $i$. Considering $f_1$: if $y_i^* = 1$, then $f_1(x_i) \ge \gamma = 1/2 + \alpha$; if $y_i^* = 0$, then $f_0(x_i) \ge \gamma$ so $f_1(x_i) \le 1-\gamma = 1/2 - \alpha$. Hence $\{x_1,\dots,x_d\}$ is $\alpha$-fat-shattered by $\mathcal{G}$ at thresholds $s_i = 1/2$, giving $\mathrm{PVC}_\gamma(\mathcal{F}) \le \mathrm{fat}_\alpha(\mathcal{G})$.
\end{proof}

Proposition~\ref{thm:sample-complexity-pvc} follows immediately from Lemma~\ref{lem:pvc-fat}. We emphasize once more that this is a \emph{lower bound} on $\mathrm{fat}_\alpha(\mathcal{G})$ in terms of $\mathrm{PVC}_\gamma(\mathcal{F})$, so it supports the positive-margin interpretation of the $\gamma$-constraint but does not give a sample-complexity bound in $\mathrm{PVC}_\gamma$. A formal uniform-convergence statement for $\mathcal{G}$ requires a separate \emph{upper} bound $\mathrm{fat}_\alpha(\mathcal{G}) \le D$, as noted above.

\section{Proof of Theorem \ref{thm:cpvc-generalization}}
\label{app:proof-theorem-1}

We prove Theorem~\ref{thm:cpvc-generalization} by applying standard uniform-convergence results separately to the induced correctness class $\mathcal{S}$ and the induced calibration-residual class $\mathcal{R}$, and combining them via a union bound. The role of C-PVC shattering is clarified at the end of the section.

\textbf{Standard regularity assumptions.}
\begin{enumerate}
    \item \textbf{Boundedness:} For any $(f, \hat{p}) \in \mathcal{F}$, the confidence function $\hat{p}_f(q) \in [0,1]$ and $s_f(z) \in \{0,1\}$; hence $r_{f,\hat{p}}(z) \in [-1,1]$.
    \item \textbf{Measurability:} All functions are measurable with respect to the product measure on $z = (C, q, J(q))$.
\end{enumerate}

\begin{proof}[Proof of Theorem \ref{thm:cpvc-generalization}]
Let $S = \{z_i\}_{i=1}^m$ be drawn i.i.d.\ from $\mathcal{D}$, with $z = (C, q, J(q))$. For each $(f,\hat{p}) \in \mathcal{F}$, recall
\[
s_f(z) = \mathbf{1}\{\mathsf{Select}_f(q) = J(q)\}, \qquad r_{f,\hat{p}}(z) = \hat{p}_f(q) - s_f(z),
\]
and the induced classes $\mathcal{S} = \{s_f : (f,\hat{p}) \in \mathcal{F}\}$, $\mathcal{R} = \{r_{f,\hat{p}} : (f,\hat{p}) \in \mathcal{F}\}$. By hypothesis, $\mathrm{VC}(\mathcal{S}) \le d_s$ and $\mathrm{fat}_\alpha(\mathcal{R}) \le d_r$ with $\alpha = \gamma - \tau - \tfrac{1}{2} > 0$. Write $D_{\gamma,\tau} = d_s + d_r$.

\textbf{Step 1: Uniform convergence for $\mathcal{S}$.}
Since each $s \in \mathcal{S}$ is $\{0,1\}$-valued and $\mathrm{VC}(\mathcal{S}) \le d_s$, the standard VC uniform-convergence inequality~\citep{anthony2009neural} gives a universal constant $K_1 > 0$ such that for any $\epsilon,\delta \in (0,1)$, whenever
\[
m \ge \frac{K_1}{\epsilon^2}\big(d_s + \log(2/\delta)\big),
\]
it holds with probability at least $1 - \delta/2$ that
\[
\sup_{s \in \mathcal{S}} \bigg| \frac{1}{m}\sum_{i=1}^m s(z_i) - \mathbb{E}[s(z)] \bigg| \le \epsilon.
\]

\textbf{Step 2: Uniform convergence for $\mathcal{R}$.}
Each $r \in \mathcal{R}$ is bounded in $[-1,1]$ and $\mathrm{fat}_\alpha(\mathcal{R}) \le d_r$ with $\alpha > 0$. The standard fat-shattering uniform-convergence bound for bounded real-valued classes~\citep{bartlett1995more, colomboni2025improved} gives a universal constant $K_2 > 0$ such that, whenever
\[
m \ge \frac{K_2}{\epsilon^2}\left(d_r \log^2(1/\epsilon) + d_r \log(1/\alpha) + \log(2/\delta)\right),
\]
it holds with probability at least $1 - \delta/2$ that
\[
\sup_{r \in \mathcal{R}} \bigg| \frac{1}{m}\sum_{i=1}^m r(z_i) - \mathbb{E}[r(z)] \bigg| \le \epsilon.
\]

\textbf{Step 3: Union bound.}
Taking a union bound over the two events and choosing the larger of the two sample-size thresholds gives both inequalities simultaneously with probability at least $1 - \delta$. The combined requirement is $m \ge \tilde{O}((D_{\gamma,\tau} + \log(1/\delta))/\epsilon^2)$, matching the compact form in Theorem~\ref{thm:cpvc-generalization} after absorbing the logarithmic factors in $1/\epsilon$ and $1/\alpha$ into the $\tilde{O}(\cdot)$ notation.

\textbf{Role of the PM condition.} The condition $\gamma > \tau + \tfrac{1}{2}$ is exactly $\alpha > 0$, which makes Step 2 applicable; outside this region the margin is non-positive and the fat-shattering bound does not apply. The theorem thus motivates the PM restriction: PM-VUS is reported precisely on the region where Step 2 is well-defined.
\end{proof}

\paragraph{Motivation lemma: positive-margin interpretation of C-PVC shattering.}
The following lemma provides a structural link between the idealized C-PVC shattering condition and the PM condition $\gamma > \tau + 1/2$. It is \emph{not} a step in the proof of Theorem~\ref{thm:cpvc-generalization}: it produces a \emph{lower bound} on a particular auxiliary class, and lower bounds do not control uniform convergence. We include it only to make the positive-margin intuition explicit.

\begin{lemma}[Positive-margin interpretation of C-PVC constraints]
\label{lemma:signed-fat-shattering}
Let $\mathcal{F}$ be a class of predictor-confidence pairs with $\mathrm{C\text{-}PVC}_{\gamma}^{\tau}(\mathcal{F}) = d_{\gamma,\tau} < \infty$ and $\gamma > \tau + \tfrac{1}{2}$. Consider the auxiliary signed-confidence family indexed by a labeling $y^* \in \{0,1\}^{|\mathcal{X}|}$:
\[
g_{f,\hat{p},y^*}(x) = (2y^*(x) - 1) \cdot \left(\hat{p}(x) - \tfrac{1}{2}\right).
\]
Then the associated class $\mathcal{G}_{\mathrm{signed}} = \{ g_{f,\hat{p},y^*} : (f, \hat{p}) \in \mathcal{F},\ y^* \in \{0,1\}^{|\mathcal{X}|}\}$ satisfies
\[
\mathrm{fat}_\alpha(\mathcal{G}_{\mathrm{signed}}) \ge d_{\gamma,\tau}
\]
at scale $\alpha = \gamma - \tau - \tfrac{1}{2} > 0$ with thresholds $s_i = 0$.
\end{lemma}

\begin{proof}
Let $S = \{x_1, \dots, x_d\}$ be a set that is $(\gamma, \tau)$-shattered under C-PVC. For any sign pattern $\sigma \in \{-1, +1\}^d$, define $y_i^* = \tfrac{1}{2}(\sigma_i + 1) \in \{0, 1\}$. By C-PVC shattering, there exists $(f, \hat{p}) \in \mathcal{F}$ such that for all $i \in [d]$: $\mathbb{P}(f(x_i) = y_i^*) \ge \gamma$ and $|\hat{p}(x_i) - \mathbb{P}(f(x_i) = y_i^*)| \le \tau$.

\textbf{Case 1 ($\sigma_i = +1$, i.e.\ $y_i^* = 1$).} Then $\hat{p}(x_i) \ge \gamma - \tau$, so $g(x_i) = \hat{p}(x_i) - \tfrac{1}{2} \ge \alpha$.

\textbf{Case 2 ($\sigma_i = -1$, i.e.\ $y_i^* = 0$).} Then $\mathbb{P}(f(x_i)=1) \le 1-\gamma$ and $\hat{p}(x_i) \le 1-\gamma+\tau$, so $g(x_i) = -(\hat{p}(x_i) - \tfrac{1}{2}) \ge \alpha$.

In both cases $\sigma_i g(x_i) \ge \alpha$ for the specific $g \in \mathcal{G}_{\mathrm{signed}}$ chosen for sign pattern $\sigma$. Since this holds for every $\sigma \in \{-1,+1\}^d$, $S$ is $\alpha$-fat-shattered at thresholds $s_i = 0$.
\end{proof}

\begin{remark}[Scope of Lemma~\ref{lemma:signed-fat-shattering}]
\label{remark:asymmetry}
This lemma does not provide a complexity upper bound. It only shows that the PM condition $\gamma > \tau + 1/2$ corresponds to a positive signed margin under the idealized C-PVC constraints. Treating the true labeling $y^*$ as part of the class index also inflates $\mathcal{G}_{\mathrm{signed}}$ beyond predictor-only variability, which is another reason it cannot be used directly as a generalization parameter. The role of Theorem~\ref{thm:cpvc-generalization} is therefore to supply a complexity envelope on \emph{actual} induced classes ($\mathcal{S}$ and $\mathcal{R}$), and the PM condition is what aligns that envelope with a positive fat-shattering margin.
\end{remark}

\begin{proof}[Proof of Theorem \ref{thm:cpvc-generalization}]
Let $S = \{z_i\}_{i=1}^m$ be drawn i.i.d.\ from $\mathcal{D}$. For each $(f,\hat{p}) \in \mathcal{F}$, recall $s_f(z) = \mathbf{1}\{\mathsf{Select}_f(q) = J(q)\}$ and $r_{f,\hat{p}}(z) = \hat{p}_f(q) - s_f(z)$, with induced classes $\mathcal{S}$ and $\mathcal{R}$ as in the statement. By hypothesis, $\mathrm{VC}(\mathcal{S}) \le d_s$ and $\mathrm{fat}_\alpha(\mathcal{R}) \le d_r$ with $\alpha > 0$.

\textbf{Step 1: Uniform convergence for $\mathcal{S}$.} Each $s_f$ is $\{0,1\}$-valued and $\mathrm{VC}(\mathcal{S}) \le d_s$. The standard VC uniform-convergence inequality~\citep{anthony2009neural} gives a universal constant $K_1 > 0$ such that, whenever $m \ge K_1 \epsilon^{-2} (d_s + \log(2/\delta))$,
\[
\sup_{s \in \mathcal{S}} \bigg| \tfrac{1}{m}\textstyle\sum_{i} s(z_i) - \mathbb{E}[s(z)] \bigg| \le \epsilon
\]
with probability at least $1 - \delta/2$.

\textbf{Step 2: Uniform convergence for $\mathcal{R}$.} Each $r_{f,\hat{p}}$ is bounded in $[-1,1]$ and $\mathrm{fat}_\alpha(\mathcal{R}) \le d_r$ with $\alpha > 0$. The standard fat-shattering uniform-convergence bound for bounded real-valued classes~\citep{bartlett1995more, colomboni2025improved} gives a universal constant $K_2 > 0$ such that, whenever $m \ge K_2 \epsilon^{-2} ( d_r \log^2(1/\epsilon) + d_r \log(1/\alpha) + \log(2/\delta))$,
\[
\sup_{r \in \mathcal{R}} \bigg| \tfrac{1}{m}\textstyle\sum_{i} r(z_i) - \mathbb{E}[r(z)] \bigg| \le \epsilon
\]
with probability at least $1 - \delta/2$.

\textbf{Step 3: Union bound.} Choose $K = \max(K_1, K_2)$. Taking a union bound over the two events yields both inequalities simultaneously with probability at least $1 - \delta$. Absorbing the logarithmic factors into the $\tilde{O}(\cdot)$ notation gives the compact form stated in Theorem~\ref{thm:cpvc-generalization}.

\textbf{Role of the PM condition.} The condition $\gamma > \tau + \tfrac{1}{2}$ is exactly $\alpha > 0$, which is what makes Step 2 applicable; outside this region the margin is non-positive and the fat-shattering bound does not apply. The theorem thus motivates the PM restriction: PM-VUS is reported precisely on the region where Step 2 is well-defined.
\end{proof}

\section{Derivation of PM-VUS Normalization}\label{appendix:pg-vus-derivation}

We explain the normalization factor in the Positive-Margin VUS (PM-VUS) definition and verify its consistency properties.

\textbf{Definition Recap.} The PM-VUS is defined as:
\begin{equation}
M\text{-}\mathrm{PM\text{-}VUS}(\mathcal{F}) = 2 \iint_{\mathcal{G}^*} \frac{1}{\gamma - 1/2} M(\gamma, \tau, \mathcal{F}) \, d\gamma \, d\tau,
\end{equation}
where $\mathcal{G}^* = \{(\gamma, \tau) : \gamma > \tau + \tfrac{1}{2}, \gamma \in (\tfrac{1}{2}, 1], \tau \in [0, \tfrac{1}{2})\}$ is the provable generalization region.

\textbf{Purpose of the Factor $\frac{1}{\gamma - 1/2}$.} For each fixed $\gamma$, the valid range of $\tau$ is $[0, \gamma - \tfrac{1}{2})$, which has length $\gamma - \tfrac{1}{2}$. The factor $\frac{1}{\gamma - 1/2}$ normalizes the $\tau$-integration at each $\gamma$, ensuring that the contribution from each $\gamma$ slice is independent of the varying $\tau$-range length, and that for $\tau$-independent metrics like PVC, the double integral reduces consistently to a single integral over $\gamma$.

\textbf{Consistency Verification for PVC.} Since PVC depends only on $\gamma$ (not $\tau$), we verify:
\begin{align*}
\mathrm{PVC}\text{-}\mathrm{PM\text{-}VUS} &= 2 \int_{1/2}^{1} \int_{0}^{\gamma - 1/2} \frac{1}{\gamma - 1/2} \mathrm{PVC}_\gamma \, d\tau \, d\gamma \\
&= 2 \int_{1/2}^{1} \mathrm{PVC}_\gamma \cdot \frac{1}{\gamma - 1/2} \cdot (\gamma - \tfrac{1}{2}) \, d\gamma = 2 \int_{1/2}^{1} \mathrm{PVC}_\gamma \, d\gamma.
\end{align*}
Thus, for PVC, the double integral correctly reduces to twice the integral over $\gamma \in (\tfrac{1}{2}, 1]$.

\textbf{Purpose of the Leading Factor $2$.} After applying the $\frac{1}{\gamma - 1/2}$ normalization, the effective $\gamma$-integration domain is $(\tfrac{1}{2}, 1]$ with length $\tfrac{1}{2}$. The factor $2$ scales this to match the full $\gamma$-range of length $1$, making PM-VUS directly comparable with full-region VUS.

\textbf{Interpretation.} This normalization ensures fair comparison: a model with uniform performance across the PM region will have PM-VUS comparable to its full-region VUS. The gap between them reflects performance degradation in the theoretically guaranteed regime.

\paragraph{Unweighted PM aggregation.} To check whether the $1/(\gamma-1/2)$ weighting drives the main ranking, we also compute the unweighted PM integral,
\[
M\text{-}\mathrm{PM\text{-}VUS}^{\mathrm{unw}}(\mathcal{F}) \;=\; \iint_{\mathcal{G}^*} M(\gamma,\tau,\mathcal{F})\,d\gamma\,d\tau,
\]
and rank models by PM-C-PVC-VUS$^{\mathrm{unw}}$. For the eleven 7--8B models in Table~\ref{tab:model_analysis_multi}, the unweighted ordering is identical on the top-3 / bottom-3 and differs only in the middle tier by at most one position. In particular, the ranking reversal in \S\ref{subsec:ranking_reversal} (JiuZhang3.0-7B near-zero; Qwen2.5-32B-Instruct $\approx 11\times$ higher) is preserved. The weighting therefore affects absolute numerical values, not the deployment-relevant conclusion.

\section{Calibration Metrics}\label{appendix:ece}

We define the Category-based Expected Calibration Error (ECE) as the weighted average of category-level calibration errors:
\begin{align}
\text{ECE} = \sum_{C \in \mathcal{C}} \frac{|C|}{N} \left| \mathbb{E}_{q \sim \mathcal{D}_C}[\hat{p}(q)] - \mathrm{SEA}(f, C) \right|,
\end{align}
where $\mathcal{C}$ is the set of categories, $|C|$ is the sample size per category, and $N$ is the total number of samples. The Actual Error (AE) measures the global failure rate:
\begin{align}
\text{AE} = \sum_{C \in \mathcal{C}} \frac{|C|}{N} (1 - \mathrm{SEA}(f, C)).
\end{align}

\textbf{Why category-based ECE?} Standard ECE groups predictions by confidence bins. Our category-based variant groups by semantic domains, aligning with our theoretical premise that shattering occurs at the category level. This directly links the global ECE metric to the local constraints in C-PVC.

\textbf{ECE vs.\ C-PVC.} While both assess calibration, they operate differently. C-PVC is a \emph{hard constraint}: a category contributes only if it strictly satisfies tolerance $\tau$. ECE is a \emph{soft average} of deviations across all categories. Consequently, low ECE is necessary but not sufficient for high C-PVC.

To illustrate: consider two equally weighted categories where one is perfectly calibrated (error $= 0$) and another has error $0.14$ with tolerance $\tau = 0.1$. The weighted ECE is $0.07$, typically considered excellent. However, under C-PVC, the second category violates the tolerance and fails to contribute, effectively halving the dimension despite low ECE. This demonstrates that ECE can mask domain-specific failures through averaging, whereas C-PVC demands uniform reliability across all shattered categories.

\section{Interpretation of SC-VUS}\label{appendix:sc-interpretation}

SC-VUS is a \emph{relative verification-cost index}, not an absolute sample-size bound. Given an assumed capacity envelope $D_{\gamma,\tau}$ on the induced classes (as in Theorem~\ref{thm:cpvc-generalization}), the uniform-convergence inequality takes the form
\[
m \ge \tilde{O}\!\left( \frac{D_{\gamma,\tau} + \log(1/\delta)}{\epsilon^2} \right),
\]
which is a function of $(\gamma,\tau)$ through $D_{\gamma,\tau}$. Because $D_{\gamma,\tau}$ is not directly observable from black-box evaluation, SC-VUS instead aggregates the empirical quantities reported at each $(\gamma,\tau)$ into a summary index that is comparable \emph{across} models under a fixed aggregation rule.

\textbf{Practical use.} Models with higher SC-VUS require proportionally more validation under the same aggregation rule than models with lower SC-VUS; no absolute sample-size claim is implied. This index should be read alongside the PVC--C-PVC gap (which quantifies the empirical cost of imposing calibration constraints) and PM-C-PVC-VUS (which reports calibrated capacity in the PM region where positive-margin uniform convergence applies).

\textbf{Cross-domain behavior.} Our cross-domain analysis suggests that models with smaller PVC--C-PVC gaps (e.g.\ JiuZhang3.0-7B) tend to have more stable calibration-constrained behavior across domains, although this stability does not translate into large PM-C-PVC-VUS. We therefore interpret SC-VUS as a comparative reliability-cost summary rather than a guarantee-generating quantity.

\section{Category-level Measurement for PVC and C-PVC}
\label{sec:category_level_pvc}

We compute PVC and C-PVC at the level of problem categories (e.g., Algebra, Geometry) rather than individual questions. This section justifies this design choice.

\textbf{Statistical tractability.} Estimating VC-style capacity at the question level is computationally intractable due to exponential growth of binary labelings ($2^n$). By defining the input space $\mathcal{X}$ as the set of semantic categories, we reduce the hypothesis space to a manageable finite set while preserving semantic diversity. Each category acts as a ``macro-input,'' enabling efficient shattering computation.

\textbf{Alignment with model organization.} LLMs acquire reasoning skills at the domain level before mastering specific problem templates. Measuring shattering over category partitions aligns with this inductive bias, requiring models to succeed across related problems rather than overfitting to single inputs.

\textbf{Stability in calibration estimation.} Individual prediction confidences are noisy. We estimate correctness probability for category $x_i$ via expectation over the category-specific distribution $\mathcal{D}_{x_i}$:
\[
\left| \mathbb{E}_{q \sim \mathcal{D}_{x_i}}[\hat{p}(q)] - \mathrm{SEA}(f, x_i) \right| \le \tau.
\]
This averaging mitigates single-question variance, providing stable signals for the C-PVC constraint.

\textbf{Theoretical consistency.} This formulation remains consistent with Definition~\ref{def:pvc}. Mapping input $x_i$ to category $C$, the condition $\mathbb{P}(f(x_i) = y_i^*) \ge \gamma$ translates to:
\[
\mathrm{SEA}(f, x_i) = \mathbb{E}_{q \sim \mathcal{D}_{x_i}} \left[ \mathbf{1}\{ \mathsf{Select}_f(q) = \mathsf{Judge}(q) \} \right] \ge \gamma.
\]
Rather than demanding uniform correctness on every question (overly restrictive and noisy), this evaluates whether the model reliably solves domain $x_i$ with probability at least $\gamma$, a valid and robust estimate of expressive capacity.

\section{On the Finiteness of the PVC/C-PVC Proxies}
\label{sec:finite_capacity_discussion}

This appendix remarks on why the empirical PVC/C-PVC proxies are always finite in our setting. This is a \emph{finiteness} statement, not a capacity-envelope upper bound on the induced classes required by Theorem~\ref{thm:cpvc-generalization}.

\textbf{Finiteness via category-level aggregation.} Measuring PVC/C-PVC at the category level restricts the effective domain to a finite partition $\{\mathcal{D}_1,\dots,\mathcal{D}_M\}$, which immediately yields
\[
\widehat{\mathrm{PVC}}_{\gamma}(\mathcal{F}) \le M, \qquad \widehat{\mathrm{C\text{-}PVC}}_{\gamma}^{\tau}(\mathcal{F}) \le M.
\]
This is a bound on the count of categories satisfying the corresponding constraints, not on any idealized VC/fat-shattering dimension of the underlying predictor class.

\textbf{Relation to classical finite-capacity results.}~\citet{anthony2009neural} (Theorems 14.18--14.19) establish finite fat-shattering capacity for bounded-weight, Lipschitz feedforward networks. For modern Transformers with residual pathways, softmax attention, and layer normalization, direct capacity analysis is nontrivial and is not attempted here. Our framework does not depend on such architecture-specific bounds: the empirical proxies are finite by construction through the category-level partition, and the role of Theorem~\ref{thm:cpvc-generalization} is to supply a uniform-convergence motivation under an \emph{assumed} capacity envelope on the induced classes, not to derive that envelope from architectural assumptions.

\FloatBarrier
\section{Per-Dataset Breakdown of the Main Diagnostic Table}\label{appendix:per_dataset_table}

Table~\ref{tab:model_analysis_per_dataset} reports the per-dataset (Math-360 / TruthfulQA / CSQA) breakdown underlying the cross-dataset aggregate in Table~\ref{tab:model_analysis_multi}, together with the exploratory PM-SC-VUS verification-cost proxy (baseline $283.5$). The PM-SC-VUS column is an exploratory verification-cost proxy and is not used in any main claim (interpretation in Appendix~\ref{appendix:sc-interpretation}).

\begin{table}[!ht]
\centering
\small
\setlength{\tabcolsep}{4pt}
\renewcommand{\arraystretch}{1.2}
\resizebox{\linewidth}{!}{
\begin{tabular}{@{}l@{\hspace{6pt}}ccc@{\hspace{8pt}}ccc@{\hspace{8pt}}c@{\hspace{8pt}}c@{}}
\toprule
& \multicolumn{3}{c}{\cellcolor{blue!6}\textbf{Full Region}}
& \multicolumn{3}{c}{\cellcolor{orange!6}\textbf{Positive-Margin (PM) Region} ($\gamma > \tau + 0.5$)}
& \cellcolor{green!6}\textbf{Calibration}
& \cellcolor{purple!6}\textbf{Accuracy} \\
\cmidrule(lr){2-4} \cmidrule(lr){5-7} \cmidrule(lr){8-8} \cmidrule(lr){9-9}
\textbf{Model}
& \textbf{PVC-VUS} $\uparrow$ & \textbf{C-PVC-VUS} $\uparrow$ & \textbf{Gap} $\downarrow$
& \textbf{PM-PVC-VUS} $\uparrow$ & \textbf{PM-C-PVC-VUS} $\uparrow$ & \textbf{PM-SC-VUS} $\uparrow$
& \textbf{CalibError} $\downarrow$
& \textbf{SEA} $\uparrow$ \\
\midrule
\midrule
\multirow{2}{*}{Qwen2.5-7B (Pretrain) \citep{yang2024qwen2}}
& 5.26 & 3.76 & \underline{1.50} & 1.25 & 0.06 & 289.77 & 0.313 & 0.556 \\
& \scriptsize(4.78/5.66/5.73) & \scriptsize(3.27/3.93/4.09) & \scriptsize(1.51/1.73/1.64) & \scriptsize(1.26/1.15/1.33) & \scriptsize(0.00/0.01/\textbf{0.18}) & \scriptsize(283.5/284.5/\textbf{301.3}) & \scriptsize(0.320/0.318/0.303) & \scriptsize(0.597/0.565/0.571) \\
\midrule
\multirow{2}{*}{Qwen2.5-7B-Instruct (SFT+RL) \citep{yang2024qwen2}}
& \underline{5.61} & \underline{4.08} & 1.53 & \underline{1.78} & \textbf{0.17} & \textbf{300.30} & 0.291 & \underline{0.599} \\
& \scriptsize(4.65/\underline{6.13}/\underline{6.04}) & \scriptsize(3.13/\textbf{4.73}/4.40) & \scriptsize(1.52/\underline{1.40}/1.64) & \scriptsize(1.24/\underline{2.07}/\underline{2.03}) & \scriptsize(0.00/\textbf{0.47}/0.03) & \scriptsize(283.5/\textbf{330.6}/286.7) & \scriptsize(0.337/\textbf{0.247}/0.290) & \scriptsize(0.581/\underline{0.613}/\underline{0.604}) \\
\midrule
\multirow{2}{*}{\shortstack[l]{Qwen2.5-Math-7B-Instruct\\(SFT+RL) \citep{yang2024qwen25mathtechnicalreportmathematical}}}
& 4.70 & 3.20 & \underline{1.50} & 0.65 & 0.002 & 283.76 & 0.335 & 0.506 \\
& \scriptsize(4.32/4.63/5.16) & \scriptsize(3.10/2.95/3.56) & \scriptsize(\underline{1.22}/1.68/\underline{1.60}) & \scriptsize(0.59/0.54/0.82) & \scriptsize(0.00/0.00/0.01) & \scriptsize(283.5/283.5/284.2) & \scriptsize(0.289/0.388/0.329) & \scriptsize(0.539/0.463/0.517) \\
\midrule
\multirow{2}{*}{Llama-3.1-8B-Instruct (SFT+DPO) \citep{meta2024llama}}
& 4.55 & 2.92 & 1.63 & 0.60 & 0.00 & 283.53 & 0.385 & 0.495 \\
& \scriptsize(\underline{4.84}/4.13/4.68) & \scriptsize(3.51/2.44/2.80) & \scriptsize(1.33/1.69/1.88) & \scriptsize(\underline{1.47}/0.34/0.00) & \scriptsize(0.00/0.00/0.00) & \scriptsize(283.5/283.5/283.5) & \scriptsize(\underline{0.284}/0.462/0.408) & \scriptsize(\underline{0.606}/0.413/0.467) \\
\midrule
\multirow{2}{*}{OpenThinker2-7B (SFT) \citep{openthoughts2023openthinker}}
& 5.37 & 3.50 & 1.87 & 1.34 & 0.002 & 283.70 & 0.357 & 0.580 \\
& \scriptsize(\textbf{5.19}/5.46/5.46) & \scriptsize(\underline{3.68}/3.36/3.47) & \scriptsize(1.51/2.10/1.99) & \scriptsize(\textbf{2.04}/0.90/1.09) & \scriptsize(\underline{0.005}/0.00/0.00) & \scriptsize(\underline{284.0}/283.5/283.5) & \scriptsize(0.298/0.395/0.378) & \scriptsize(\textbf{0.649}/0.544/0.546) \\
\midrule
\multirow{2}{*}{\shortstack[l]{DeepSeek-R1-Distill-Qwen-7B\\(Distill) \citep{guo2025deepseek}}}
& 5.10 & 3.49 & 1.61 & 1.19 & 0.07 & 290.49 & 0.339 & 0.549 \\
& \scriptsize(4.77/5.01/5.51) & \scriptsize(3.24/3.40/3.83) & \scriptsize(1.53/1.61/1.68) & \scriptsize(1.26/1.00/1.31) & \scriptsize(0.00/0.07/\underline{0.13}) & \scriptsize(283.5/291.0/296.9) & \scriptsize(0.329/0.358/0.330) & \scriptsize(0.596/0.500/0.550) \\
\midrule
\multirow{2}{*}{Bespoke-Stratos-7B (SFT) \citep{bespokelabs2023stratos}}
& 5.33 & 3.54 & 1.79 & 1.17 & 0.00 & 283.53 & 0.347 & 0.570 \\
& \scriptsize(4.44/5.67/5.89) & \scriptsize(2.73/3.95/3.94) & \scriptsize(1.71/1.72/1.95) & \scriptsize(0.68/1.39/1.44) & \scriptsize(0.00/0.00/0.00) & \scriptsize(283.5/283.5/283.5) & \scriptsize(0.388/0.315/0.338) & \scriptsize(0.556/0.567/0.588) \\
\midrule
\multirow{2}{*}{JiuZhang3.0-7B (SFT) \citep{zhou2024jiuzhang}}
& 4.79 & 3.84 & \textbf{0.95} & 0.49 & 0.009 & 284.47 & \textbf{0.209} & 0.516 \\
& \scriptsize(4.49/4.55/5.34) & \scriptsize(\textbf{3.74}/3.22/\underline{4.57}) & \scriptsize(\textbf{0.75}/\textbf{1.33}/\textbf{0.77}) & \scriptsize(0.72/0.08/0.68) & \scriptsize(\textbf{0.02}/0.00/0.008) & \scriptsize(\textbf{285.6}/283.5/284.3) & \scriptsize(\textbf{0.169}/0.309/\textbf{0.149}) & \scriptsize(0.561/0.454/0.533) \\
\midrule
\multirow{2}{*}{\shortstack[l]{Ministral-8B-Instruct-2410\\(SFT+RL) \citep{mistralai2024ministral}}}
& 4.88 & 3.00 & 1.88 & 0.73 & 0.00 & 283.53 & 0.402 & 0.522 \\
& \scriptsize(4.21/4.61/5.81) & \scriptsize(2.50/2.65/3.85) & \scriptsize(1.71/1.96/1.96) & \scriptsize(0.53/0.38/1.28) & \scriptsize(0.00/0.00/0.00) & \scriptsize(283.5/283.5/283.5) & \scriptsize(0.414/0.453/0.339) & \scriptsize(0.528/0.458/0.579) \\
\midrule
\multirow{2}{*}{Open-Reasoner-Zero-7B (RL) \citep{hu2025openreasonerzeroopensourceapproach}}
& 5.29 & 3.51 & 1.78 & 1.19 & 0.02 & 285.60 & 0.352 & 0.566 \\
& \scriptsize(4.52/5.47/5.88) & \scriptsize(2.78/3.71/4.02) & \scriptsize(1.74/1.76/1.86) & \scriptsize(0.83/1.13/1.60) & \scriptsize(0.00/0.06/0.00) & \scriptsize(283.5/289.7/283.5) & \scriptsize(0.388/0.341/0.327) & \scriptsize(0.564/0.546/0.588) \\
\midrule
\multirow{2}{*}{s1.1-7B (SFT) \citep{muennighoff2025s1simpletesttimescaling}}
& \textbf{5.83} & \textbf{4.27} & 1.56 & \textbf{2.08} & \underline{0.12} & \underline{295.46} & \underline{0.285} & \textbf{0.622} \\
& \scriptsize(4.76/\textbf{6.22}/\textbf{6.52}) & \scriptsize(3.22/\underline{4.71}/\textbf{4.88}) & \scriptsize(1.54/1.51/1.64) & \scriptsize(1.25/\textbf{2.17}/\textbf{2.81}) & \scriptsize(0.00/\underline{0.22}/\underline{0.13}) & \scriptsize(283.5/\underline{305.8}/\underline{297.0}) & \scriptsize(0.331/\textbf{0.256}/\underline{0.267}) & \scriptsize(0.593/\textbf{0.621}/\textbf{0.652}) \\
\bottomrule
\end{tabular}
}
\caption{\textbf{Per-dataset breakdown of Table~\ref{tab:model_analysis_multi}.} First row per model: cross-dataset aggregate (identical to Table~\ref{tab:model_analysis_multi}). Second row (parentheses): Math-360 / TruthfulQA / CSQA. PM-SC-VUS is an exploratory verification-cost proxy (baseline $283.5$; Appendix~\ref{appendix:sc-interpretation}) and is not used in any main claim. \textbf{Bold}: best; \underline{underline}: second best.}
\label{tab:model_analysis_per_dataset}
\end{table}
\FloatBarrier

\section{Evaluation on MATH-500 Benchmark}\label{appendix:math500}

To assess robustness, we replicate our evaluation on \textsc{MATH-500} \citep{huggingface2024math500}. Table~\ref{tab:math500_analysis} presents the results.

\begin{table}[!ht]
\centering
\small
\setlength{\tabcolsep}{4pt}
\renewcommand{\arraystretch}{1.2}
\resizebox{\linewidth}{!}{
\begin{tabular}{@{}l@{\hspace{6pt}}ccc@{\hspace{8pt}}ccc@{\hspace{8pt}}c@{\hspace{8pt}}c@{}}
\toprule
& \multicolumn{3}{c}{\cellcolor{blue!6}\textbf{Full Region}}
& \multicolumn{3}{c}{\cellcolor{orange!6}\textbf{Positive-Margin (PM) Region} ($\gamma > \tau + 0.5$)}
& \cellcolor{green!6}\textbf{Calibration} 
& \cellcolor{purple!6}\textbf{Accuracy} \\
\cmidrule(lr){2-4} \cmidrule(lr){5-7} \cmidrule(lr){8-8} \cmidrule(lr){9-9}
\textbf{Model} 
& \textbf{PVC-VUS} $\uparrow$ & \textbf{C-PVC-VUS} $\uparrow$ & \textbf{Gap} $\downarrow$ 
& \textbf{PM-PVC-VUS} $\uparrow$ & \textbf{PM-C-PVC-VUS} $\uparrow$ & \textbf{PM-SC-VUS} $\uparrow$ 
& \textbf{CalibError} $\downarrow$
& \textbf{SEA} $\uparrow$ \\
\midrule
\midrule
Qwen2.5-7B (Pretrain) \citep{yang2024qwen2} 
& 3.93 & 2.56 & 1.37 & 0.71 & 0.00 & 283.5 & 0.356 & 0.561 \\
\midrule
Qwen2.5-7B-Instruct (SFT+RL) \citep{yang2024qwen2} 
& 3.94 & 2.54 & 1.40 & 0.69 & 0.00 & 283.5 & 0.358 & 0.562 \\
\midrule
\shortstack[l]{Qwen2.5-Math-7B-Instruct\\(SFT+RL) \citep{yang2024qwen25mathtechnicalreportmathematical}} 
& 3.68 & 2.57 & \underline{1.11} & 0.20 & 0.00 & 283.5 & 0.304 & 0.524 \\
\midrule
Llama-3.1-8B-Instruct (SFT+DPO) \citep{meta2024llama} 
& 3.97 & 2.67 & 1.30 & 0.75 & 0.00 & 283.5 & 0.331 & 0.565 \\
\midrule
OpenThinker2-7B (SFT) \citep{openthoughts2023openthinker} 
& \underline{4.36} & \underline{3.09} & 1.27 & \underline{1.43} & 0.00 & 283.5 & \underline{0.294} & \underline{0.622} \\
\midrule
\shortstack[l]{DeepSeek-R1-Distill-Qwen-7B\\(Distill) \citep{guo2025deepseek}} 
& 3.95 & 2.52 & 1.43 & 0.67 & 0.00 & 283.5 & 0.369 & 0.563 \\
\midrule
Bespoke-Stratos-7B (SFT) \citep{bespokelabs2023stratos} 
& 4.19 & 2.88 & 1.31 & 1.24 & \textbf{0.11} & \textbf{294.1} & 0.330 & 0.597 \\
\midrule
JiuZhang3.0-7B (SFT) \citep{zhou2024jiuzhang} 
& 3.74 & \textbf{3.11} & \textbf{0.63} & 0.35 & 0.00 & 283.5 & \textbf{0.169} & 0.534 \\
\midrule
\shortstack[l]{Ministral-8B-Instruct-2410\\(SFT+RL) \citep{mistralai2024ministral}} 
& 3.97 & 2.50 & 1.47 & 0.81 & 0.00 & 283.5 & 0.377 & 0.566 \\
\midrule
Open-Reasoner-Zero-7B (RL) \citep{hu2025openreasonerzeroopensourceapproach} 
& \textbf{4.39} & 2.99 & 1.40 & \textbf{1.50} & 0.00 & 283.5 & 0.322 & \textbf{0.626} \\
\midrule
s1.1-7B (SFT) \citep{muennighoff2025s1simpletesttimescaling} 
& 4.26 & 2.93 & 1.33 & 1.25 & 0.00 & 283.5 & 0.317 & 0.607 \\
\bottomrule
\end{tabular}
}
\caption{Self-evaluation performance on MATH-500 benchmark. Results are consistent with the main findings (Table~\ref{tab:model_analysis_multi}): the expressivity--calibration trade-off persists, with JiuZhang3.0-7B achieving the lowest Gap ($0.63$) and best calibration. Notably, PM-C-PVC-VUS is zero for $10$ of $11$ models; only Bespoke-Stratos-7B achieves nonzero calibration-constrained behavior ($0.11$) in the positive-margin region, indicating even more severe calibration limitations on this benchmark. \textbf{Bold}: best; \underline{underline}: second best.}
\label{tab:math500_analysis}
\end{table}
\FloatBarrier

\begin{figure*}[!b]
\centering
\includegraphics[width=\textwidth]{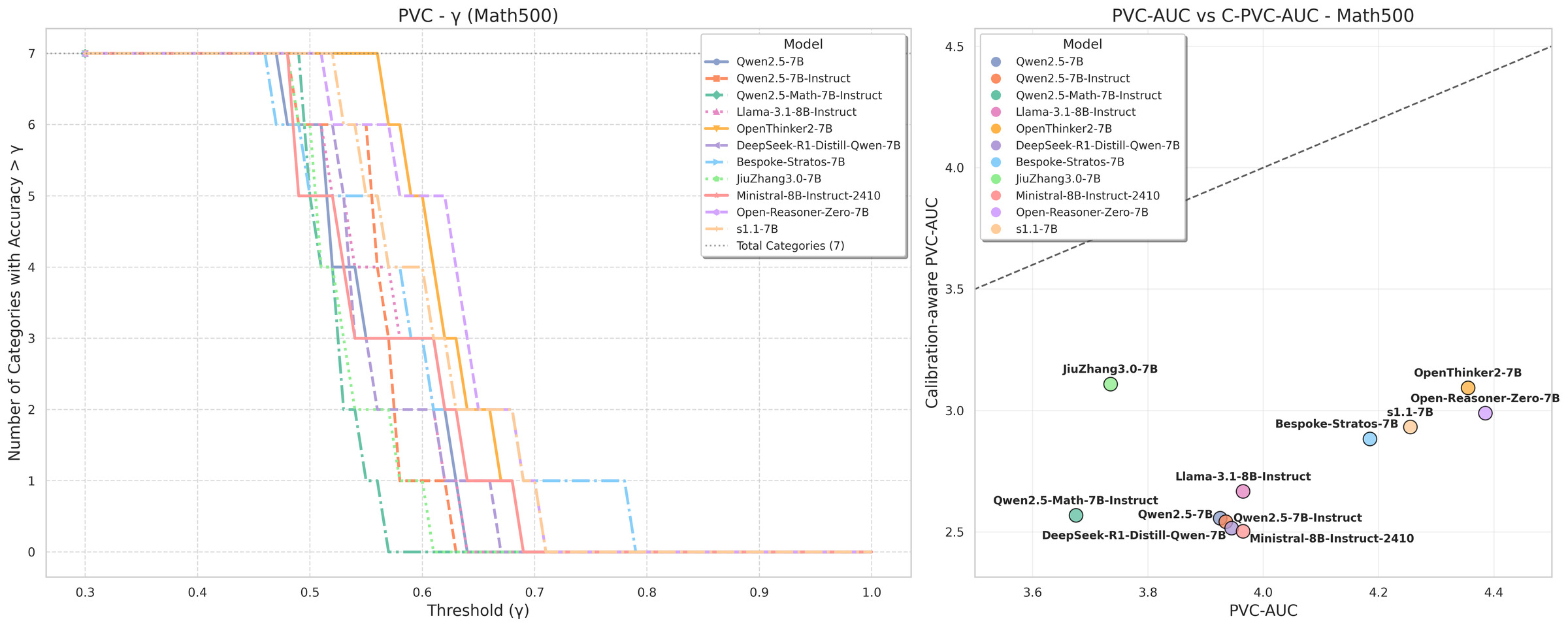}
\caption{
(a) PVC scores across \revision{accuracy thresholds} ($\gamma$).
(b) Comparison between PVC-VUS and C-PVC-VUS. JiuZhang3.0-7B shows the smallest gap, demonstrating superior calibration.
}
\label{fig:combined_calibration_pvc_plot_math500_overall}
\end{figure*}

\begin{figure*}[ht]
\centering
\includegraphics[width=\textwidth]{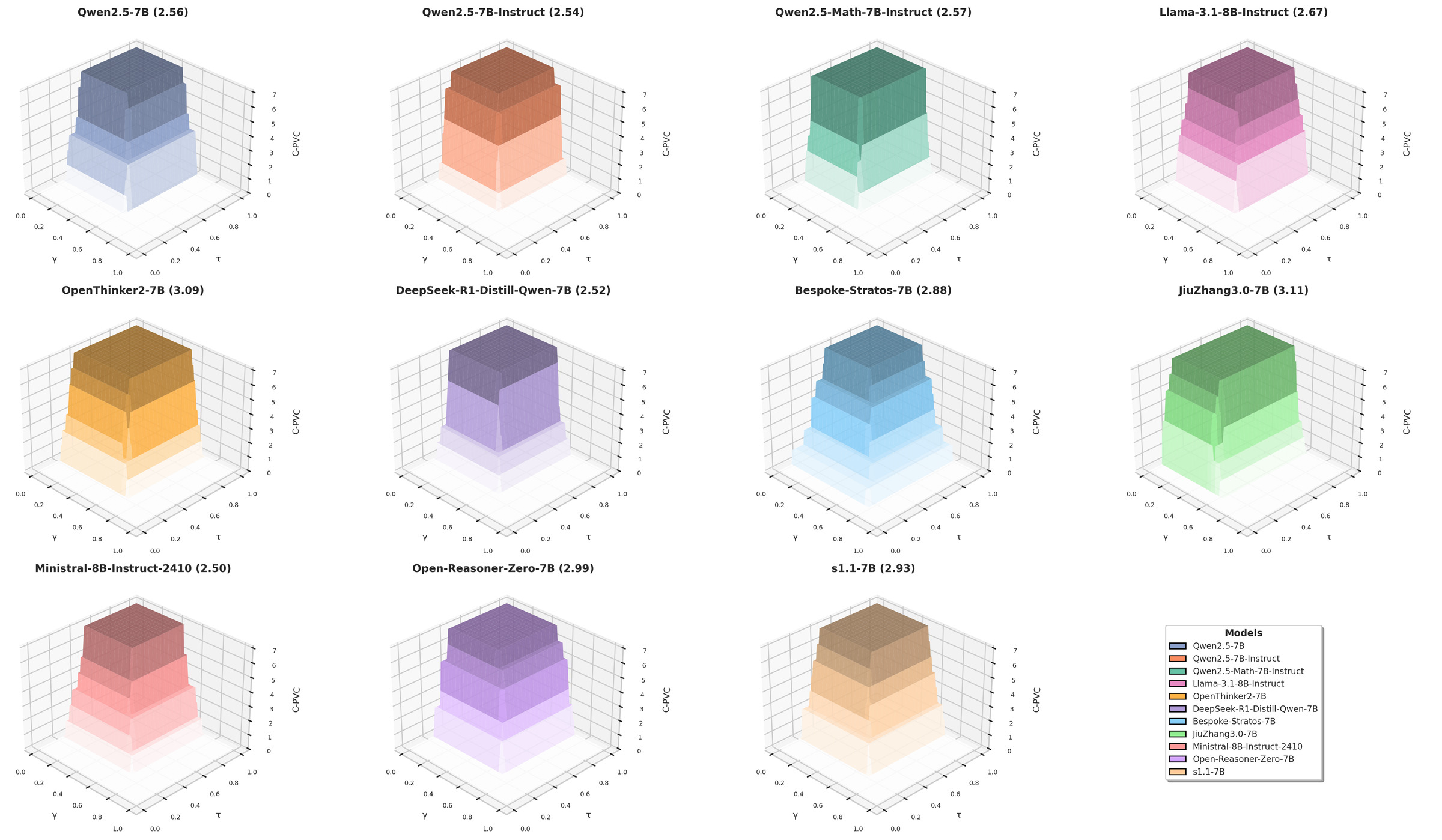}
\caption{
Empirical C-PVC diagnostic surfaces on MATH-500. The x-axis ($\tau$) represents calibration tolerance, y-axis ($\gamma$) shows the accuracy threshold, and z-axis displays the empirical C-PVC proxy.
}
\label{fig:all_models_cpvc_3d_grid_math500}
\end{figure*}

The findings largely mirror our main evaluation, reinforcing the robustness of our framework. JiuZhang3.0-7B maintains superior calibration, achieving the lowest Gap (0.63) and CalibError (0.169). Open-Reasoner-Zero-7B and OpenThinker2-7B lead in expressiveness (PVC-VUS: 4.39 and 4.36 respectively), consistent with their strong performance on Math-360.

Critically, PM-C-PVC-VUS equals zero for 10 of 11 models; only Bespoke-Stratos-7B achieves nonzero calibrated capacity (0.11) in the provable generalization region. This is notably worse than Math-360 results, where multiple models achieved nonzero PM-C-PVC-VUS, suggesting that reliable self-evaluation becomes more challenging on this benchmark.

Some ranking deviations point to benchmark-specific effects. Llama-3.1-8B-Instruct shows degraded calibration on MATH-500 compared to Math-360, despite maintaining moderate PVC-VUS. These cross-dataset results validate our framework's stability while highlighting that calibration behavior can vary across problem distributions.

\FloatBarrier
\section{Extended Cross-Domain Analysis}
\label{appendix:cross_domain_vis}

Extended domain-specific signatures (main-text Figure~\ref{fig:combined_model_performance}): OpenThinker2-7B achieves the highest Math-360 performance (PVC-VUS $=5.19$, SEA $=64.9\%$); Qwen2.5-7B-Instruct peaks on TruthfulQA (PVC-VUS $=6.13$); s1.1-7B leads on CSQA (PVC-VUS $=6.52$) with consistently high performance across domains. Calibration is more cross-domain consistent than raw discrimination: JiuZhang3.0-7B maintains the smallest PVC--C-PVC gap across Math-360, TruthfulQA, and CSQA ($0.75$, $1.33$, $0.77$), suggesting calibration behavior generalizes more robustly across domain boundaries than discrimination. Main-text Figure~\ref{fig:combined_calibration_pvc_plot} visualizes the threshold and cross-calibration signatures.

\FloatBarrier
\section{Judge Correlation Study}\label{sec:judge_correlation}

\begin{figure*}[ht]
\centering
\includegraphics[width=\textwidth]{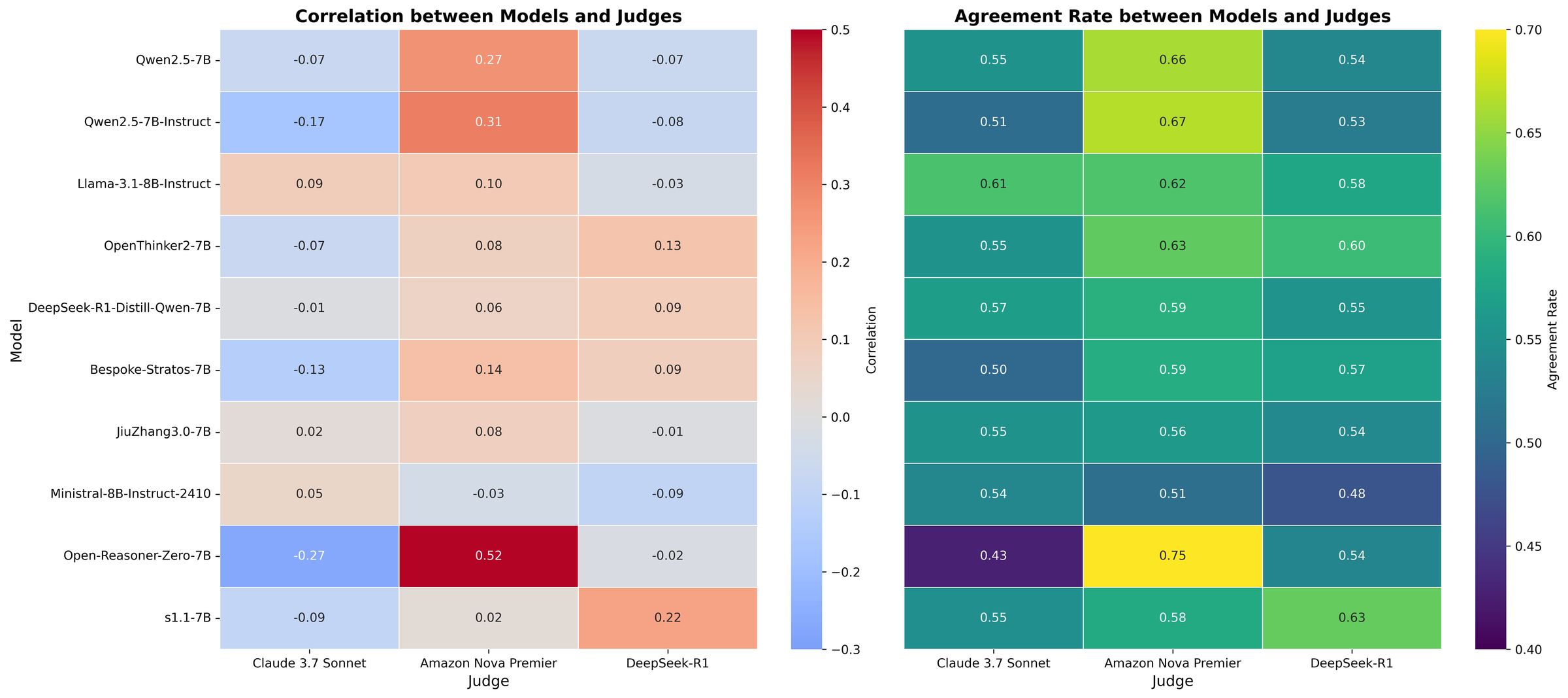}
\caption{Judge correlation analysis. Left: Pearson correlation between model self-evaluation and judge verdicts. Right: Direct agreement rates. Judges exhibit distinct evaluative patterns, justifying our ensemble approach.}
\label{fig:model_judge_comparison}
\end{figure*}

To assess reference-label consistency, we analyzed correlations between model self-assessments and verdicts from three reference LLMs (Claude 3.7 Sonnet, Amazon Nova Premier, DeepSeek-R1) on Math-360. Figure~\ref{fig:model_judge_comparison} shows Pearson correlations and agreement rates; Table~\ref{tab:overall_correlation} summarizes overall statistics.

\textbf{Judge characteristics.} The three judges exhibit distinct evaluative patterns. Amazon Nova Premier shows the highest alignment with target models (average correlation: 0.18, agreement: 0.61), suggesting evaluation criteria well-matched to 7--8B model reasoning patterns. Claude 3.7 Sonnet exhibits negligible or negative correlations (average: $-$0.09, agreement: 0.53), indicating stricter or more conservative evaluation standards. DeepSeek-R1 occupies middle ground (average correlation: 0.01, agreement: 0.55), providing balanced assessment without strong bias toward particular model families.

\textbf{Justification for ensemble.} The divergence in evaluative strictness substantiates our multi-judge protocol. Reliance on a single judge risks systematic bias: Claude's strictness may underestimate capabilities while Nova's alignment may overestimate them. Notably, the overall accuracy of models often exceeds agreement with any individual judge, indicating that single judges have non-overlapping error modes. The majority-consensus rule integrates these complementary perspectives, reducing sensitivity to individual judge preferences while we do not claim the resulting reference labels are oracle annotations.

\begin{table}[h!]
\centering
\begin{tabular}{l c c}
\toprule
\textbf{LLM Judge} & \textbf{Correlation} & \textbf{Agreement} \\
\midrule
Claude 3.7 Sonnet & $-$0.09 & 0.53 \\
Amazon Nova Premier & 0.18 & 0.61 \\
DeepSeek-R1 & 0.01 & 0.55 \\
\bottomrule
\end{tabular}
\caption{Overall Pearson correlation and agreement rate between judges and model self-assessments on Math-360.}
\label{tab:overall_correlation}
\end{table}

\FloatBarrier
\section{Reference-Label Robustness: Unanimous-Only and Label-Perturbation Analyses}
\label{appendix:robustness}

This appendix provides the detailed reference-label robustness analyses summarized in \S\ref{subsec:robustness}. Both analyses are designed to test whether the near-zero PM-C-PVC-VUS regime is a judge-noise artifact.

\subsection{Unanimous-only re-analysis}
\label{appendix:unanimous}

We restrict the evaluation to pairs on which all three LLM judges agree (3-0 unanimous vote). This removes contested cases at the cost of a smaller effective sample size, and serves as a cleaner-label stress test. Table~\ref{tab:unanimous_pgcpvc} reports PM-C-PVC-VUS under the full-label and unanimous-only settings.

\begin{table}[h]
\centering
\small
\renewcommand{\arraystretch}{1.15}
\setlength{\tabcolsep}{6pt}
\begin{tabular}{@{}lcc@{}}
\toprule
\textbf{Model} & \textbf{PM-C-PVC-VUS (full)} & \textbf{PM-C-PVC-VUS (unanimous)} \\
\midrule
s1.1-7B                & 0.142 & \textbf{0.996} \\
Qwen2.5-32B-Instruct   & 0.209 & 0.597 \\
Claude 3.5 Haiku       & 0.123 & 0.588 \\
JiuZhang3.0-7B         & 0.019 & 0.047 \\
Llama-3.1-8B-Instruct  & 0.000 & 0.038 \\
\bottomrule
\end{tabular}
\caption{PM-C-PVC-VUS under full-label vs.\ unanimous-only re-analysis. Restricting to 3-0 unanimous judge cases substantially raises PM-C-PVC-VUS for stronger models (e.g.\ s1.1-7B to $0.996$), but the ranking structure and the near-zero regime for weaker models (JiuZhang3.0-7B, Llama-3.1-8B-Instruct) are preserved. The central pattern is therefore not an artifact of 2-1 judge splits.}
\label{tab:unanimous_pgcpvc}
\end{table}

Restricting to unanimous cases raises mean PM-C-PVC-VUS from $0.073$ (full) to $0.320$ (unanimous) across the five models shown, with a wider spread ($0.047$ to $0.996$) that preserves the original ranking structure. In particular, JiuZhang3.0-7B stays near zero ($0.019 \to 0.047$), indicating that its calibration-constrained deficit is not driven by judge disagreement.

\subsection{Synthetic judge-label perturbation}
\label{appendix:label-perturbation}

We further test whether the near-zero PM-C-PVC-VUS regime is a label-noise artifact by introducing synthetic perturbation: each reference label is independently flipped with probability $p \in \{0.10, 0.20, 0.30\}$. We repeat this $5$ times per perturbation level with independent seeds and report mean $\pm$ one standard deviation. Table~\ref{tab:label_perturbation} reports PVC-VUS and PM-C-PVC-VUS for two representative models: s1.1-7B (best-in-class for raw discrimination at the 7B tier) and JiuZhang3.0-7B (near-zero PM-C-PVC-VUS baseline), under $0\%$, $10\%$, and $30\%$ flip probability.

\begin{table}[h]
\centering
\small
\renewcommand{\arraystretch}{1.15}
\setlength{\tabcolsep}{6pt}
\begin{tabular}{@{}llcc@{}}
\toprule
\textbf{Model} & \textbf{Flip} & \textbf{PVC-VUS} & \textbf{PM-C-PVC-VUS} \\
\midrule
s1.1-7B          & 0\%  & 20.90           & 0.142 \\
s1.1-7B          & 10\% & 20.09 $\pm$ 0.30 & 0.103 $\pm$ 0.053 \\
s1.1-7B          & 30\% & 18.42 $\pm$ 0.58 & 0.059 $\pm$ 0.064 \\
\midrule
JiuZhang3.0-7B   & 0\%  & 17.40           & 0.019 \\
JiuZhang3.0-7B   & 10\% & 17.22 $\pm$ 0.18 & 0.019 $\pm$ 0.014 \\
JiuZhang3.0-7B   & 30\% & 17.03 $\pm$ 0.29 & 0.023 $\pm$ 0.026 \\
\bottomrule
\end{tabular}
\caption{Label-perturbation robustness: PVC-VUS and PM-C-PVC-VUS under independent random flips of $10$--$30\%$ of reference labels (mean $\pm$ one standard deviation over $5$ seeds per level). PVC-VUS degrades gradually with noise as expected, but the near-zero PM-C-PVC-VUS regime for JiuZhang3.0-7B remains stable under $30\%$ corruption ($0.019$--$0.023$). The central finding is therefore not primarily a judge-noise artifact.}
\label{tab:label_perturbation}
\end{table}

Two observations are worth highlighting. First, raw discrimination (PVC-VUS) degrades smoothly with label noise, as expected from standard concentration: s1.1-7B drops from $20.90$ to $18.42$ at $30\%$ flip probability. Second, PM-C-PVC-VUS for models already near zero under clean labels stays near zero under noise: JiuZhang3.0-7B's PM-C-PVC-VUS moves only from $0.019$ to $0.019$--$0.023$ under $10$--$30\%$ corruption. A label-noise explanation would predict that flipping labels should either substantially depress high-PM models toward zero (which occurs gradually) or inflate low-PM models away from zero (which does not occur). The observed behavior is consistent with the finding that calibration-constrained capacity in the positive-margin region is \emph{genuinely scarce}, not an artifact of noisy reference labels.

\FloatBarrier
\section{Decoupling of Reasoning and Self-Evaluation}
\label{appendix:independence_analysis}

A fundamental question in the study of metacognition is whether self-evaluation capabilities are decoupled from primary reasoning skills, or merely a byproduct of generation performance. To investigate this, we analyzed the correlation between \textbf{Solution Accuracy} (SA), the proportion of correctly solved problems, and \textbf{Self-Evaluation Accuracy} (SEA), the proportion of correct pairwise judgments, at the category level across all 11 models and four benchmarks.

\begin{figure*}[h]
\centering
\includegraphics[width=\textwidth]{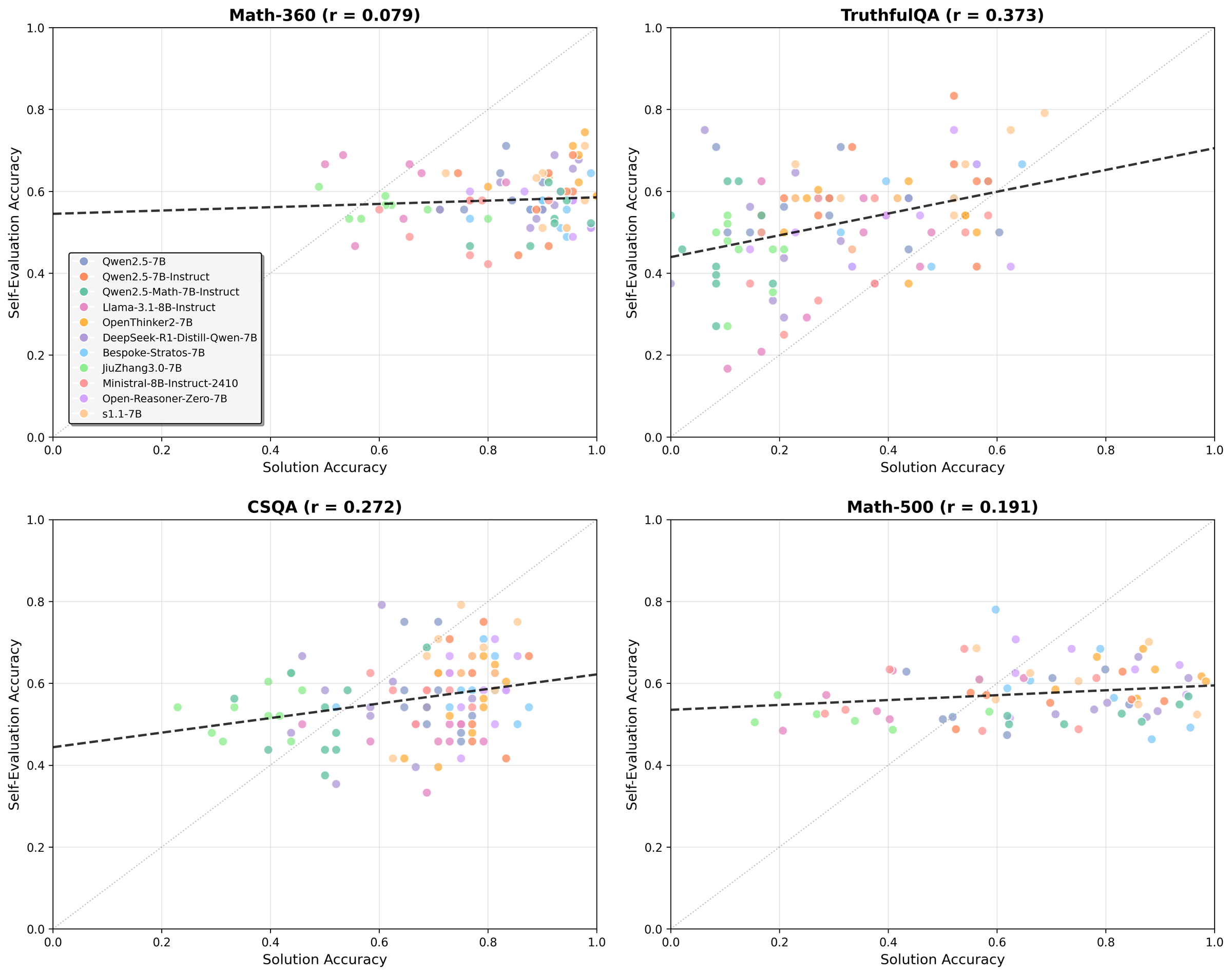}
\caption{Correlation between Solution Accuracy (x-axis) and Self-Evaluation Accuracy (y-axis) across four benchmarks. Each point represents a (model, category) pair. The consistently low Pearson correlation coefficients ($r$) across all datasets indicate that generation and evaluation capabilities are largely independent.}
\label{fig:acc_vs_sea_correlation}
\end{figure*}

\subsection{Weak Correlation Across Domains}

As illustrated in Figure~\ref{fig:acc_vs_sea_correlation}, we observe consistently weak correlations between generation correctness and evaluation capability across all datasets. This dissociation is particularly pronounced in mathematical reasoning: the Pearson correlation is negligible on Math-360 ($r=0.079$) and remains weak on MATH-500 ($r=0.191$). Even on TruthfulQA ($r=0.148$) and CSQA ($r=0.186$), correlations remain modest. This statistical independence suggests that a model's proficiency in solving problems does \emph{not} guarantee its capacity to discriminate between superior and inferior solutions.

\subsection{Inverse Performance Patterns}

Table~\ref{tab:acc_vs_sea_data} provides quantitative evidence of this decoupling through specific \emph{inverse patterns}: cases where models with lower reasoning performance achieve higher evaluation accuracy.

\textbf{Case 1: MATH-500.} Open-Reasoner-Zero-7B achieves a Solution Accuracy of 76.6\%, substantially lower than OpenThinker2-7B's 86.7\% (a gap of 10.1 percentage points). Yet, Open-Reasoner-Zero-7B exhibits \emph{higher} Self-Evaluation Accuracy (62.6\% vs.\ 62.2\%). If self-evaluation were simply a function of reasoning ability, we would expect the stronger reasoner to also be the better evaluator.

\textbf{Case 2: TruthfulQA.} s1.1-7B records a lower SA (44.2\%) than Bespoke-Stratos-7B (45.2\%), yet demonstrates vastly superior SEA (62.1\% vs.\ 56.7\%, a gap of 5.4 percentage points). This inversion is particularly striking given the similar reasoning performance.

\textbf{Case 3: Math-360.} Llama-3.1-8B-Instruct achieves only 63.9\% SA compared to Qwen2.5-7B-Instruct's 89.6\%, yet attains \emph{higher} SEA (60.6\% vs.\ 58.1\%). Despite a 25.7 percentage point deficit in problem-solving, Llama demonstrates superior self-assessment.

\begin{table*}[t]
\centering
\vspace{1mm}
\small
\renewcommand{\arraystretch}{1.1}
\setlength{\tabcolsep}{5pt}
\begin{tabular}{@{}lcccccccc@{}}
\toprule
\multirow{2}{*}{\textbf{Model}} 
& \multicolumn{2}{c}{\textbf{Math-360}} 
& \multicolumn{2}{c}{\textbf{TruthfulQA}} 
& \multicolumn{2}{c}{\textbf{CSQA}} 
& \multicolumn{2}{c}{\textbf{MATH-500}} \\
\cmidrule(lr){2-3} \cmidrule(lr){4-5} \cmidrule(lr){6-7} \cmidrule(lr){8-9}
& SA & SEA & SA & SEA & SA & SEA & SA & SEA \\
\midrule
Qwen2.5-7B              & 83.1 & 59.7 & 29.2 & 56.5 & 70.0 & 57.1 & 63.0 & 56.1 \\
Qwen2.5-7B-Instruct     & 89.6 & 58.1 & 40.2 & 61.3 & 78.7 & 60.4 & 70.6 & 56.2 \\
Qwen2.5-Math-7B-Instruct & 90.8 & 53.9 & 9.4  & 46.3 & 49.4 & 51.7 & 79.2 & 52.4 \\
Llama-3.1-8B-Instruct   & 63.9 & 60.6 & 30.4 & 41.3 & 68.5 & 46.7 & 41.4 & 56.5 \\
OpenThinker2-7B         & \textbf{94.7} & \textbf{64.9} & 39.0 & 54.4 & 75.8 & 54.6 & \textbf{86.7} & 62.2 \\
DeepSeek-R1-Distill     & 91.8 & 59.6 & 17.5 & 50.0 & 57.5 & 55.0 & 83.9 & 56.3 \\
Bespoke-Stratos-7B      & 90.7 & 55.6 & \textbf{45.2} & 56.7 & 77.3 & 58.8 & 76.0 & 59.7 \\
JiuZhang3.0-7B          & 61.7 & 56.1 & 13.8 & 45.4 & 37.1 & 53.3 & 36.0 & 53.4 \\
Ministral-8B-Instruct   & 75.7 & 52.8 & 33.3 & 45.8 & 68.8 & 57.9 & 52.2 & 56.6 \\
Open-Reasoner-Zero-7B   & 91.5 & 56.4 & 39.4 & 54.6 & \textbf{79.0} & 58.8 & 76.6 & \textbf{62.6} \\
s1.1-7B                 & 90.7 & 59.3 & 44.2 & \textbf{62.1} & 75.6 & \textbf{65.2} & 75.4 & 60.7 \\
\bottomrule
\end{tabular}
\caption{Comparison of Solution Accuracy (SA) vs.\ Self-Evaluation Accuracy (SEA) across four benchmarks. All values in percentages (\%). Notable inverse patterns, where models with lower SA achieve higher SEA (e.g., Open-Reasoner-Zero vs.\ OpenThinker2 on MATH-500; Llama-3.1 vs.\ Qwen2.5-Instruct on Math-360), demonstrate that self-evaluation is not merely a byproduct of reasoning capability. \textbf{Bold}: best per metric per dataset.}
\label{tab:acc_vs_sea_data}
\end{table*}

\subsection{Implications for Self-Evaluation Assessment}

These findings carry important implications for our framework:

\textbf{Self-evaluation as an orthogonal skill.} The weak correlations and inverse patterns collectively suggest that self-evaluation operates as a \emph{distinct metacognitive capability}, structurally independent of raw reasoning generation. A model may excel at solving problems while being poor at recognizing correct solutions, or vice versa.

\textbf{Necessity of specialized metrics.} Standard accuracy benchmarks capture only generative proficiency (SA), leaving evaluative reliability (SEA) unmeasured. Our PVC and C-PVC dimensions provide orthogonal insights into model trustworthiness that cannot be inferred from task performance alone.

\textbf{Practical deployment considerations.} For autonomous systems that must self-verify outputs, these results caution against assuming that high task accuracy implies reliable self-assessment. A model selected purely for reasoning performance may exhibit poor introspective calibration, leading to overconfident errors in deployment.

In summary, the empirical decoupling of reasoning and self-evaluation validates the necessity of our specialized framework: metrics like C-PVC capture a fundamentally different aspect of model capability than traditional accuracy measures.

\FloatBarrier
\section{On the Insufficiency of Rank-Based Metrics for Self-Evaluation Assessment}
\label{app:auroc_vs_cpvc}

\subsection{The Isotonic Invariance of AUROC}

A natural question in the evaluation of probabilistic reasoning systems is why established discriminative metrics, specifically the Area Under the Receiver Operating Characteristic curve (AUROC), are insufficient for quantifying the reliability of self-evaluation. The fundamental limitation of AUROC in this context stems from its property of \emph{isotonic invariance}, which renders it blind to the absolute meaning of confidence scores.

Formally, let $\mathcal{D}$ be a distribution over inputs $\mathcal{X}$ and binary labels $\mathcal{Y} \in \{0, 1\}$. For a probabilistic predictor $f: \mathcal{X} \to [0, 1]$, the AUROC can be interpreted as the probability that a randomly selected positive instance $x^+$ is assigned a higher score than a randomly selected negative instance $x^-$:
\begin{equation}
    \text{AUROC}(f) = \mathbb{P}_{(x^+, x^-) \sim \mathcal{D}^2} \left( f(x^+) > f(x^-) \right).
\end{equation}
This definition depends exclusively on the \emph{ordinal ranking} of predictions, independent of their absolute magnitudes. Consequently, for any strictly monotonic transformation $g: [0, 1] \to [0, 1]$, the AUROC remains invariant:
\begin{equation}
    \text{AUROC}(f) = \text{AUROC}(g \circ f).
\end{equation}
While this invariance is advantageous for tasks requiring only relative discrimination, it is fundamentally inadequate for autonomous systems where the absolute value of $f(x)$, the reported confidence, must serve as a reliable proxy for the true probability of correctness.

\subsection{The Calibration Blindness Problem}

The isotonic invariance of AUROC leads to a critical failure mode we term \emph{Calibration Blindness}: a model may exhibit perfect discriminative power while being dangerously miscalibrated.

\textbf{Illustrative Example.} Consider two predictors performing self-evaluation on a category of problems, both achieving 65\% accuracy but differing in reported confidence. \textbf{Predictor $f_A$} (well-calibrated) reports 65\% confidence, yielding calibration error $|0.65 - 0.65| = 0$. \textbf{Predictor $f_B$} (overconfident) reports 99\% confidence, yielding calibration error $|0.99 - 0.65| = 0.34$. Any rank-based metric, including AUROC, would treat $f_A$ and $f_B$ as equivalent since their ordinal rankings are identical. However, deploying $f_B$ in high-stakes environments poses serious risks: the system would proceed based on \emph{hallucinated certainty}, bypassing uncertainty-based safety mechanisms.

\subsection{How C-PVC Addresses Calibration Blindness}

The Calibration-aware Probabilistic VC (C-PVC) dimension provides a principled remedy by introducing a metric explicitly sensitive to the \emph{absolute} values of confidence scores. For a category to contribute to the C-PVC dimension, a predictor must satisfy both accuracy threshold $\gamma$ and calibration tolerance $\tau$:
\begin{equation}
    \left| \hat{p}(x) - \mathbb{E}\!\left[ \mathbf{1}_{\{f(x)=y^*\}} \right] \right| \le \tau.
\end{equation}
Applying this constraint with $\tau = 0.1$:
\[
f_A: |0.65 - 0.65| = 0.00 \le 0.1 \ \checkmark \qquad f_B: |0.99 - 0.65| = 0.34 > 0.1 \ \text{\ding{55}}
\]
Thus, C-PVC correctly penalizes $f_B$ while accepting $f_A$, distinguishing between models that merely rank correctly versus those with trustworthy self-knowledge.

\subsection{The Critical Role of the Positive-Margin (PM) Region}

While C-PVC addresses calibration blindness, a further question arises: \emph{under what conditions do these measurements admit a uniform-convergence interpretation?} This is where the \textbf{PM region} becomes essential.

Theorem~\ref{thm:cpvc-generalization} establishes that generalization bounds for C-PVC hold only when $\gamma > \tau + \frac{1}{2}$. This constraint ensures a positive fat-shattering margin ($\alpha = \gamma - \tau - \frac{1}{2} > 0$), which is necessary for uniform convergence guarantees. The PM-VUS metrics restrict evaluation to this theoretically grounded regime:
\begin{equation}
    \mathcal{G}^* = \left\{(\gamma, \tau) : \gamma > \tau + \tfrac{1}{2}\right\}.
\end{equation}

\textbf{Why PM-C-PVC-VUS Matters.} Consider the contrast between full-region and PM-region metrics from our experiments (Table~\ref{tab:model_analysis_multi}):

\begin{center}
\small
\begin{tabular}{lcc}
\toprule
\textbf{Model} & \textbf{C-PVC-VUS} (Full) & \textbf{PM-C-PVC-VUS} \\
\midrule
Qwen2.5-7B-Instruct & 4.08 & 0.17 \\
s1.1-7B & 4.27 & 0.12 \\
JiuZhang3.0-7B & 3.84 & 0.009 \\
\bottomrule
\end{tabular}
\end{center}

The dramatic drop from C-PVC-VUS to PM-C-PVC-VUS points to a critical insight: \emph{most of the measured calibrated capacity exists outside the PM region, where the uniform-convergence motivation of Theorem~\ref{thm:cpvc-generalization} does not apply}. A model may appear well-calibrated under lenient thresholds without its calibration surviving in the positive-margin region.

\textbf{Connecting the Three Levels.} Our framework provides a hierarchical assessment of self-evaluation reliability:

\begin{enumerate}
    \item \textbf{AUROC / PVC}: Measures pure discriminative capacity: can the model rank correct answers higher? This is necessary but insufficient for trustworthy deployment.
    
    \item \textbf{C-PVC}: Adds calibration constraints: does the model's confidence align with its accuracy? This addresses calibration blindness but without the positive-margin condition.

    \item \textbf{PM-C-PVC}: Restricts to the positive-margin region: the stringent operating region where observed calibration admits a uniform-convergence interpretation under capacity-envelope assumptions. This is the most demanding level of our diagnostic hierarchy.
\end{enumerate}

\textbf{The Sobering Conclusion.} Our experiments show that even the best-performing models achieve PM-C-PVC-VUS $\approx 0.17$ (Qwen2.5-7B-Instruct), compared to PM-PVC-VUS $\approx 1.78$. This means that within the regime where generalization is provable:
\begin{itemize}
    \item Models retain \emph{moderate} discriminative capacity (PM-PVC-VUS $> 0$)
    \item Models exhibit \emph{near-zero} calibrated capacity (PM-C-PVC-VUS $\approx 0$)
\end{itemize}

In other words, even when we can guarantee that a model's discrimination ability will generalize, we \emph{cannot} guarantee that its calibration will. This fundamental asymmetry, invisible to AUROC and only partially captured by full-region C-PVC, is precisely what PM-C-PVC-VUS measures.

\subsection{Summary: A Hierarchy of Self-Evaluation Metrics}

\begin{table}[h]
\centering
\small
\begin{tabular}{lccc}
\toprule
\textbf{Metric} & \textbf{Discrimination} & \textbf{Calibration} & \textbf{Generalization} \\
\midrule
AUROC & \checkmark & \ding{55} & \ding{55} \\
PVC-VUS & \checkmark & \ding{55} & Partial \\
C-PVC-VUS & \checkmark & \checkmark & \ding{55} \\
PM-PVC-VUS & \checkmark & \ding{55} & \checkmark \\
\textbf{PM-C-PVC-VUS} & \checkmark & \checkmark & \checkmark \\
\bottomrule
\end{tabular}
\caption{Hierarchy of self-evaluation metrics and their guarantees. PVC-VUS provides partial generalization bounds that depend on the accuracy threshold $\gamma$ but lack calibration constraints.}\label{tab:metric_hierarchy}
\end{table}

Table~\ref{tab:metric_hierarchy} summarizes the progression from rank-based metrics to our full framework. The PM-C-PVC-VUS metric represents the most stringent test of self-evaluation reliability: it requires discrimination, calibration, \emph{and} provable generalization. The near-zero values observed across all models (best: $0.17$ for Qwen2.5-7B-Instruct) underscore our central finding: \textbf{reliable, generalizable self-evaluation remains largely unattained in current LLMs}.

\FloatBarrier
\section{Comparative Advantages of the PVC Framework}
\label{app:pvc_advantages}

While standard metrics such as Expected Calibration Error (ECE) and Area Under the ROC Curve (AUROC) provide valuable insights into model performance, they exhibit fundamental limitations when assessing the structural reliability of self-evaluation. The Probabilistic VC (PVC) framework addresses these shortcomings through two key properties: calibration awareness and categorical decomposability.

\subsection{Beyond Ranking and Averages}

As detailed in Appendix~\ref{app:auroc_vs_cpvc}, rank-based metrics like AUROC suffer from isotonic invariance, rendering them blind to calibration errors. Similarly, ECE provides only a global average where opposing errors across domains can cancel out, obscuring domain-specific failures. For instance, a model that is overconfident in mathematics but underconfident in factual reasoning may achieve low ECE while being unreliable in both domains individually.

In contrast, C-PVC imposes \emph{per-category} calibration constraints: each domain must independently satisfy the tolerance $\tau$, preventing error cancellation and ensuring uniform reliability across all evaluated categories.

\subsection{Categorical Decomposability}

A practical advantage of the PVC framework is its natural alignment with categorical evaluation. When the input space admits a disjoint partition into semantic domains $\mathcal{X}_{\text{total}} = \bigcup_{i} \mathcal{X}_i$ (e.g., $\mathcal{X}_{\text{Math}}$, $\mathcal{X}_{\text{Logic}}$, $\mathcal{X}_{\text{Fact}}$), our empirical PVC measurement, defined as the count of shattered categories, is inherently additive:
\begin{equation}
    \begin{aligned}
        \widehat{\text{PVC}}_{\gamma} &= \sum_{i} \mathbf{1}\{\text{SEA}(\mathcal{X}_i) \geq \gamma\} \\
        \widehat{\text{C-PVC}}_{\gamma}^{\tau} &= \sum_{i} \mathbf{1}\{\text{SEA}(\mathcal{X}_i) \geq \gamma \text{ and } \text{CalibError}(\mathcal{X}_i) \leq \tau\}
    \end{aligned}
\end{equation}

This formulation offers two practical benefits. First, \textbf{independent evaluation}: each category's contribution is determined solely by its own accuracy and calibration, unaffected by other categories. Adding or removing a category does not change the shattered status of existing categories. Second, \textbf{fine-grained diagnostics}: the per-category breakdown directly shows which domains contribute to (or detract from) overall capacity, enabling targeted analysis of model strengths and weaknesses.

We note that this additivity holds for our \emph{empirical} category-count measurement at fixed $(\gamma, \tau)$. When aggregating via VUS, comparisons should be made across evaluations with the same category set to ensure consistency.

\subsection{Summary of Comparative Advantages}

Table~\ref{tab:metric_comparison} summarizes these distinctions. Unlike AUROC (discrimination only) or ECE (global calibration with cancellation), C-PVC combines discrimination, per-category calibration constraints, decomposability, and, via the PM restriction, a positive-margin uniform-convergence motivation, into a single diagnostic framework.

\begin{table}[h]
\centering
\small
\begin{tabular}{lccc}
\toprule
\textbf{Property} & \textbf{AUROC} & \textbf{ECE} & \textbf{C-PVC} \\
\midrule
Discrimination-aware & \checkmark & \ding{55} & \checkmark \\
Calibration-aware & \ding{55} & \checkmark & \checkmark \\
Per-category constraints & \ding{55} & \ding{55} & \checkmark \\
Decomposable & \ding{55} & Partial & \checkmark \\
Generalization bounds & \ding{55} & \ding{55} & \checkmark \\
\bottomrule
\end{tabular}
\caption{Comparison of self-evaluation metrics across key properties.}
\label{tab:metric_comparison}
\end{table}

\FloatBarrier
\section{Detailed Experimental Methodology} 
\label{sec:detailed-experimental-methodology}

This section elaborates on our empirical approach to measuring PVC dimensions in language models, providing implementation details for the framework introduced in Section~\ref{subsec:measuring_pvc_cpvc}.

\subsection{Measurement Protocol}

Our empirical evaluation follows a three-stage protocol designed to measure a model's capacity for reasoning self-assessment while avoiding information leakage. The process is illustrated in Figure~\ref{fig:exp-protocol}.

\paragraph{Stage~1: Solution Generation.}
For each problem instance, the model generates two independent solutions using distinct decoding trajectories: one following standard step-by-step reasoning patterns, and another exploring a non-standard, potentially more intuitive or compact reasoning path. These variants ensure diversity in reasoning quality while remaining entirely model-generated.

\paragraph{Stage~2: Self-Evaluation.}
The model is presented with both candidate solutions in randomized order and must select the answer it believes to be more correct while assigning a confidence score to its choice. Crucially, the model is \emph{not} given access to the reference label, ensuring that this choice reflects genuine internal evaluation rather than supervised feedback.

\paragraph{Stage~3: Reference Labeling.}
To assign the reference preference, a separate three-judge LLM ensemble evaluates the same solution pair with access to the correct final answer. The ensemble assigns the reference label via majority vote. This design keeps the reference signal independent of the model being evaluated and reduces sensitivity to individual judge biases.

\subsection{Hyperparameters}
\label{sec:hyperparameters}

Table~\ref{tab:decode_judge_config} summarizes the decoding and evaluation configuration applied identically to all eleven models: Qwen2.5-7B, Qwen2.5-7B-Instruct~\citep{yang2024qwen2}, Qwen2.5-Math-7B-Instruct~\citep{yang2024qwen25mathtechnicalreportmathematical}, DeepSeek-R1-Distill-Qwen-7B~\citep{guo2025deepseek}, Llama-3.1-8B-Instruct~\citep{meta2024llama}, OpenThinker2-7B~\citep{openthoughts2023openthinker}, Bespoke-Stratos-7B~\citep{bespokelabs2023stratos}, JiuZhang3.0-7B~\citep{zhou2024jiuzhang}, Ministral-8B-Instruct-2410~\citep{mistralai2024ministral}, Open-Reasoner-Zero-7B~\citep{hu2025openreasonerzeroopensourceapproach}, and s1.1-7B~\citep{muennighoff2025s1simpletesttimescaling}. The same settings were used for all judge ensemble queries.

\begin{table}[h]
\centering
\small
\begin{tabular}{lll}
\toprule
\textbf{Stage} & \textbf{Parameter} & \textbf{Value} \\
\midrule
\multirow{4}{*}{Solution Generation} 
& Temperature & 0.7 \\
& Top-P (nucleus sampling) & 0.9 \\
& Max tokens & 4096 \\
& Solutions per problem & 2 \\
\midrule
\multirow{3}{*}{Self-Evaluation} 
& Presentation order & Randomized \\
& Confidence scale & 0--100\% \\
& Reference-label access & None \\
\midrule
\multirow{4}{*}{External Judging} 
& Judge 1 & Claude 3.7 Sonnet~\citep{anthropic2024claude37} \\
& Judge 2 & Amazon Nova Premier~\citep{intelligence2024amazon} \\
& Judge 3 & DeepSeek-R1~\citep{guo2025deepseek} \\
& Aggregation & Majority vote \\
\bottomrule
\end{tabular}
\caption{Experimental configuration for solution generation and evaluation.}
\label{tab:decode_judge_config}
\end{table}

\subsection{Math-360 Benchmark Construction}
\label{sec:math360}

To instantiate the diagnostic protocol on a contamination-aware mathematical setting, we developed Math-360, a controlled diagnostic testbed of $360$ original problems. Math-360 is used to support the evaluation protocol under balanced domain, difficulty, and subcategory structure, and is not positioned as a standalone general-purpose math leaderboard; its role in the paper is supporting, with cross-domain validation provided by TruthfulQA, CSQA, and MATH-500. Unlike prior math benchmarks derived from widely available competition problems or textbooks (which may overlap with LLM pretraining corpora), Math-360 emphasizes originality and balanced domain coverage.

\paragraph{Taxonomic Coverage.} 
Table~\ref{tab:math_taxonomy} presents the taxonomy of mathematical reasoning categories. The dataset spans eight core domains (Arithmetic, Algebra, Calculus, Geometry, Number Theory, Combinatorics, Statistics, and Linear Algebra), each subdivided into five foundational subtopics. This taxonomy reflects standard mathematics curricula and facilitates interpretable analysis across well-defined conceptual boundaries.

\begin{table*}[t]
\centering
\vspace{1mm}
\small
\setlength{\tabcolsep}{8pt}
\renewcommand{\arraystretch}{1.15}
\begin{tabular}{@{}ll@{}}
\toprule
\textbf{Category} & \textbf{Subcategories} \\
\midrule
Arithmetic      & Basic Operations, Fractions, Percentages, Numerical Approximation, Order of Operations \\
Algebra         & Equations, Inequalities, Polynomials, Functions, Systems of Equations \\
Calculus        & Differentiation, Integration, Limits, Series, Applications \\
Geometry        & Plane Geometry, Coordinate Geometry, Transformations, Mensuration, Trigonometry \\
Number Theory   & Divisibility, Modular Arithmetic, Primes, Diophantine Equations, Number Sequences \\
Combinatorics   & Counting Principles, Permutations, Combinations, Probability, Recursion \\
Statistics      & Descriptive Statistics, Distributions, Hypothesis Testing, Regression, Bayesian Inference \\
Linear Algebra  & Matrices, Determinants, Vector Spaces, Eigenvalues, Linear Transformations \\
\bottomrule
\end{tabular}
\caption{Taxonomy of mathematical reasoning categories in our benchmark.}
\label{tab:math_taxonomy}
\end{table*}

\revision{
\paragraph{Problem stratification and complexity.}
Each subcategory contains five problems rigorously stratified across difficulty levels (Easy, Medium, Hard), yielding a total of 360 problems. Crucially, while the Easy tier validates fundamental competencies, the Hard tier incorporates {complex, multi-step reasoning tasks}. This broad difficulty spectrum is designed to stress-test the reasoning limits of 7--8B models. We note that the examples in Table~\ref{tab:math_examples} are selected for brevity and illustrative clarity; the actual dataset includes problems requiring extensive intermediate derivations to prevent ceiling effects.
}

\paragraph{Representative examples.} Table~\ref{tab:math_examples} provides one representative problem per subcategory, highlighting the range and depth of reasoning skills required. Problems are designed to elicit multi-step thinking, abstraction, and formal manipulation, rather than direct recall or pattern matching. For instance, problems in combinatorics test combinatorial enumeration and recurrence, while those in number theory evaluate modular reasoning and structural properties of integers.

Taken together, the benchmark offers a principled and contamination-resistant testbed for evaluating mathematical reasoning in language models. It supports both overall capability assessment and fine-grained diagnosis of strengths and weaknesses across mathematical domains.

\begin{table*}[!h]
\centering
\resizebox{\textwidth}{!}{%
\begin{tabular}{lll}
\toprule 
\textbf{Category} & \textbf{Subcategory} & \textbf{Example Problem} \\
\midrule
\multirow{5}{*}{Arithmetic} & Basic Operations & Calculate $238 + 149$. \\
& Fractions & What is $\frac{2}{5} + \frac{1}{5}$? \\
& Percentages & What is $25\%$ of $80$? \\
& Numerical Approximation & Round $47.68$ to the nearest whole number. \\
& Order of Operations & Calculate $3 + 4 \times 2$. \\
\midrule
\multirow{5}{*}{Algebra} & Equations & Solve for $x$: $x + 5 = 12$. \\
& Inequalities & Solve: $x + 3 > 7$. \\
& Polynomials & Simplify: $(3x^2 + 2x) + (4x^2 - 5x + 1)$. \\
& Functions & If $f(x) = 2x + 3$, find $f(4)$. \\
& Systems of Equations & Solve the system: $x + y = 5, x - y = 1$. \\
\midrule
\multirow{5}{*}{Calculus} & Differentiation & Find the derivative of $f(x) = 3x^2 + 2x - 5$. \\
& Integration & Find $\int (3x^2 + 2) \, dx$. \\
& Limits & Evaluate $\lim_{x \to 3} (x^2 - 4)$. \\
& Series & Find the sum of the first 10 terms of the arithmetic sequence with $a_1 = 3$ and $d = 4$. \\
& Applications & Find the maximum value of $f(x) = -x^2 + 6x - 5$ on the interval $[0, 5]$. \\
\midrule
\multirow{5}{*}{Geometry} & Plane Geometry & Find the area of a rectangle with length $8 \text{ cm}$ and width $5 \text{ cm}$. \\
& Coordinate Geometry & Find the distance between the points $(3, 4)$ and $(6, 8)$. \\
& Transformations & Reflect the point $(3, 5)$ across the $x$-axis. \\
& Mensuration & Find the circumference of a circle with radius $5 \text{ cm}$. \\
& Trigonometry & Find $\sin(30^\circ)$. \\
\midrule
\multirow{5}{*}{Number Theory} & Divisibility & Determine whether $156$ is divisible by $4$. \\
& Modular Arithmetic & Calculate $17 \pmod 5$. \\
& Primes & List all prime numbers less than $20$. \\
& Diophantine Equations & Find all integer solutions to $x + y = 10$. \\
& Number Sequences & Find the next number in the sequence: $3, 7, 11, 15, \dots$ \\
\midrule
\multirow{5}{*}{Combinatorics} & Counting Principles & How many different 3-digit numbers can be formed using the digits $1, 2, 3, 4, 5$ without repetition? \\
& Permutations & How many permutations can be formed using all the letters of the word `MATH'? \\
& Combinations & How many ways are there to select 3 books from a shelf containing 7 different books? \\
& Probability & A fair die is rolled. What is the probability of getting a number greater than 4? \\
& Recursion & Find the 6th term in the Fibonacci sequence, where $F_1 = 1, F_2 = 1$, and $F_n = F_{n-1} + F_{n-2}$ for $n \ge 3$. \\
\midrule
\multirow{5}{*}{Statistics} & Descriptive Statistics & Find the mean of the numbers $4, 7, 9, 3, \text{ and } 2$. \\
& Distributions & If $X$ follows a normal distribution with mean $10$ and standard deviation $2$, what is $P(X > 12)$? \\
& Hypothesis Testing & In a hypothesis test, what is the meaning of a Type I error? \\
& Regression & In a linear regression model, what does the slope coefficient represent? \\
& Bayesian Inference & State Bayes' theorem. \\
\midrule
\multirow{5}{*}{Linear Algebra} & Matrices & Calculate the sum of matrices $A = \begin{bmatrix} 3 & 5 \\ 1 & 2 \end{bmatrix}$ and $B = \begin{bmatrix} 4 & 0 \\ 6 & 3 \end{bmatrix}$. \\
& Determinants & Calculate the determinant of the matrix $A = \begin{bmatrix} 3 & 4 \\ 1 & 2 \end{bmatrix}$. \\
& Vector Spaces & Determine whether the set of vectors $\{(1, 0, 0), (0, 1, 0), (0, 0, 1)\}$ forms a basis for $\mathbb{R}^3$. \\
& Eigenvalues & Find the eigenvalues of the matrix $A = \begin{bmatrix} 2 & 0 \\ 0 & 3 \end{bmatrix}$. \\
& Linear Transformations & Let $T: \mathbb{R}^2 \to \mathbb{R}^2$ be defined by $T(x, y) = (2x, 3y)$. Find $T(1, 2)$. \\
\bottomrule
\end{tabular}
}
\caption{Representative problems from each subcategory in our mathematical reasoning benchmark 360.}
\label{tab:math_examples}
\end{table*}

\subsection{Construction of Non-Mathematical Benchmarks}
\label{sec:other_dataset}

To investigate the generalizability of our framework beyond mathematical reasoning, we incorporated two established benchmarks: TruthfulQA~\citep{lin-etal-2022-truthfulqa} and CommonsenseQA~\citep{talmor-etal-2019-commonsenseqa}. However, utilizing these datasets in their raw form presents challenges for our category-level analysis due to inherent class imbalances: the original datasets contain highly uneven distributions across semantic topics, which could bias PVC/C-PVC estimates toward dominant categories.

To ensure statistical stability, we implemented a curation protocol aligned with Math-360's balanced structure. First, we identified the top-10 major categories per benchmark based on semantic labels to ensure broad domain coverage. Second, we determined the minimum sample count among these categories as a uniform balancing constraint. Third, we performed stratified sampling of exactly 24 questions from each category, yielding a standardized subset of 240 questions per benchmark ($10 \text{ categories} \times 24 \text{ questions}$). This uniform distribution ensures that PVC and C-PVC estimates reflect true capability across diverse reasoning patterns rather than being skewed by data volume in specific sub-domains.

\subsection{Experimental Prompts}

Our measurement procedure operationalizes the theoretical concept of PVC dimension through a three-stage prompting process.

\subsubsection{First Solution Generation}

The target model generates an expert-level solution using the following prompt:

\begin{promptbox}{First Solution Prompt}
\small
\textbf{System:} You are an expert mathematics educator with deep knowledge across all mathematical domains. Your task is to solve a math problem with exceptional clarity and accuracy.

\textbf{User:} Please solve the following mathematical problem step by step, ensuring complete accuracy:

PROBLEM: \{\textcolor{pinktext}{problem}\}

Provide a detailed solution that demonstrates expert-level understanding.

Include:
1. Clear explanation of your approach
2. Each step fully worked out
3. Proper mathematical notation
4. Verification of your answer

Make sure your final answer is clearly indicated and correct.
\end{promptbox}

\subsubsection{Second Solution Generation}

The model generates an alternative solution using a prompt that requests creative approaches:

\begin{promptbox}{Second Solution Prompt}
\small
\textbf{System:} You are a brilliant mathematics educator known for your creative problem-solving approaches. Your task is to solve a math problem using your new methods.

\textbf{User:} Please solve the following mathematical problem:

PROBLEM: \{\textcolor{pinktext}{problem}\}

Provide an extremely detailed solution that demonstrates understanding.

Include:
1. A clear explanation of your approach using an alternative and creative solution path
2. Each step fully worked out with mathematical reasoning
3. Proper mathematical notation and formalism
4. If applicable, connect this problem to broader mathematical concepts or real-world applications

Make your solution as thorough and educational as possible, showing multiple ways to think about and verify the solution. Focus on approaches that demonstrate mathematical insight and elegance.
\end{promptbox}

This design choice, requesting alternative approaches rather than explicitly introducing errors, allows for natural quality variation while avoiding artificially degraded solutions that would be too obviously flawed.

\subsubsection{Self-Evaluation Prompt}

The model evaluates both solutions (presented in randomized order), identifies which it believes is superior, and provides a confidence score (0--100\%):

\begin{promptbox}{Self-Evaluation Prompt}
\small
\textbf{System:} You are an expert mathematics evaluator tasked with comparing two solutions to the same problem.

\textbf{User:} Below are two different solutions to the same mathematical problem. Your task is to carefully evaluate both solutions and determine which one is superior in terms of correctness, clarity, and mathematical insight.

PROBLEM: \{\textcolor{pinktext}{problem}\}

SOLUTION A: \{\textcolor{pinktext}{solution\_a}\}

SOLUTION B: \{\textcolor{pinktext}{solution\_b}\}

Please compare these solutions carefully and respond with:
1. Which solution is better (A or B)
2. Your confidence in this judgment (0-100\%)
3. A brief explanation of why one solution is superior to the other

Format your response as:
Better solution: [A or B]
Confidence: [0-100\%]
Explanation: [Your analysis]
\end{promptbox}

The reported confidence score directly operationalizes the $\hat{p}(x)$ term in our C-PVC definition (Definition~\ref{def:c-pvc}), enabling calibration error computation at the category level.

\subsubsection{Prompts for Non-Mathematical Benchmarks}

For TruthfulQA and CommonsenseQA, we adapted the above prompts by replacing mathematical terminology with domain-appropriate language (e.g., "factual accuracy" for TruthfulQA, "commonsense reasoning" for CommonsenseQA) while preserving the three-stage structure. The self-evaluation prompt structure remains identical, ensuring consistent confidence elicitation across all domains.

\end{document}